\documentclass[journal]{IEEEtran}

\usepackage{hyperref}       
\usepackage{courier}
\usepackage{url}            
\usepackage{booktabs}       
\usepackage{amsfonts}       
\usepackage{nicefrac}       
\usepackage{microtype}      
\usepackage{amsmath}
\usepackage{bm}
\usepackage{xcolor,framed}
\usepackage{caption}
\usepackage{subcaption}
\usepackage{graphicx}
\usepackage{wrapfig,lipsum,booktabs}
\usepackage{wrapfig}
\usepackage{float}

\begin{document}

\title{\LARGE \bf
DynaNet: Neural Kalman Dynamical Model \\ for Motion Estimation and Prediction
}

%


\author{Changhao Chen, Chris Xiaoxuan Lu, Bing Wang, Niki Trigoni and Andrew Markham 
\thanks{Changhao Chen is with the College of Intelligence Science and Technology, National University of Defense Technology, Changsha, 410073, China}
\thanks{Bing Wang, Niki Trigoni and Andrew Markham are with the Department of Computer Science, University of Oxford, Oxford, OX1 3QD, United Kingdom.} 
\thanks{Chris Xiaoxuan Lu is with the School of Informatics, University of Edinburgh, Edinburgh, EH8 9AB, United Kingdom}
\thanks{Corresponding author: Changhao Chen (chenchanghao@nudt.edu.cn)}
\thanks{This work was supported by NFSC (Grant number: 62103427, 62073331) and EPSRC Program ``ACE-OPS: From Autonomy to Cognitive assistance in Emergency OPerationS" (Grant number: EP/S030832/1)}
}

\markboth{IEEE Transactions on Neural Networks and Learning Systems,~Vol.~X, No.~X, Sep~2021}%
{Shell \MakeLowercase{\textit{et al.}}: Bare Demo of IEEEtran.cls for IEEE Journals}

\maketitle

\begin{abstract}
Dynamical models estimate and predict the temporal evolution of physical systems. State Space Models (SSMs) in particular represent the system dynamics  with many desirable properties, such as being able to model uncertainty in both the model and measurements, and optimal (in the Bayesian sense)  recursive formulations e.g. the Kalman Filter. However, they require significant domain knowledge to derive the parametric form and considerable hand-tuning to correctly set all the parameters. Data driven techniques e.g. Recurrent Neural Networks have emerged as compelling alternatives to SSMs with wide success across a number of challenging  tasks, in part due to their impressive capability to extract relevant features from rich inputs. They however lack interpretability and robustness to unseen conditions. Thus, data-driven models are hard to be applied in safety-critical applications, such as self-driving vehicles. In this work, we present DynaNet, a hybrid deep learning and time-varying state-space-model which can be trained end-to-end. Our neural Kalman dynamical model allows us to exploit the relative merits of both state-space model and deep neural networks. We demonstrate its effectiveness of the estimation and prediction on a number of physically challenging tasks, including visual odometry, sensor fusion for visual-inertial navigation and motion prediction. In addition we show how DynaNet can indicate failures through investigation of properties such as the rate of innovation (Kalman Gain).
\end{abstract}


\section{Introduction}
From catching a ball to tracking the motion of planets across the celestial sphere, the ability to estimate and predict the future trajectory of moving objects is key for interaction with our physical world. With ever increasing automation e.g. self-driving vehicles and mobile robotics, the ability to not only estimate system states based on sensor data, but also to reason about latent dynamics and therefore predict states with partial or even without any observation is of huge importance to the safety and reliability of intelligent systems~\cite{Sunderhauf2018}. 

Newtonian/classical mechanics has been developed as an explicit mathematical model which can be used to predict future motion and infer how an object has moved in the past. This is commonly captured in a state-space-model (SSM) that describes the temporal relationship and evolution of states through first-order differential equations. For example, in the task of estimating egomotion from visual sensors, also known as Visual Odometry (VO) \cite{Nister2004,Engel2013,Forster2014}, velocity, position, and orientation are usually chosen as physically attributable states for mobile robots. These models are typically hand-crafted based on domain knowledge and require significant expertise to develop and tune their hyper-parameters. Simplifying assumptions are often made, for example, to treat the system as being linear, time-invariant with uncertainty being additive and Gaussian. A canonical example of an optimal Bayesian filter for linear systems is the Kalman Filter \cite{Kalman1960}, which is an optimal linear quadratic estimator. Although capable of controlling sophisticated mechanical systems (e.g. the Apollo 11 lander used a 21 state Kalman Filter \cite{Mohinder2010}), it becomes more challenging to use in complex, nonlinear systems, giving rise to alternative variants such as the Sequential Monte Carlo~\cite{Liu1998} or nonlinear graph optimisation~\cite{Kummerle2011}. However, even when using these sophisticated approaches, imperfections in model parameters and measurement errors from sensory data contribute to issues such as accumulative drift in visual navigation systems. Furthermore, there is a disconnect between the complexities of rich sensor data e.g. images and derived states.

Identifying the underlying mechanism governing the motion of an object is a hard problem to solve for dynamical systems operating in real world. As a consequence, in recent years, there has been an explosive growth in applying deep neural networks (DNNs) for motion estimation \cite{Clark2017,Wang2017,Zhou2017,Clark2017a,Mirowski2018,Bloesch2018,Brahmbhatt2018,Zhan2018,Yin2018}. These learning based approaches can extract useful representations from high-dimensional raw data to estimate the key motion states in an end-to-end fashion. 

Although these learned models demonstrate good performance in both accuracy and robustness, they are `black-boxes' regressing measurements with implicit latent states and difficult to interpret. 
In contrast to neural networks, state-space-models (SSMs) are able to construct a parametric model description and offer an explicit transition relation that describes the evolution of system states and uncertainty into the future. They can also optimally fuse measurements from multiple sensors based on their innovation gain, rather than simply stacking them as in a neural network.

\begin{figure*}
    \includegraphics[width=0.95\textwidth]{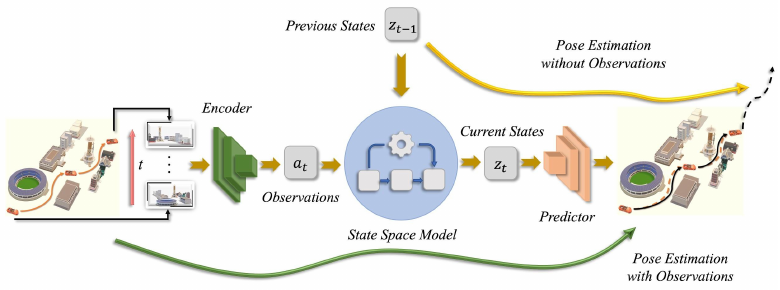}
    \captionof{figure}{A concept figure of our proposed DynaNet that combines deep neural networks (DNNs) and state-space-models (SSMs) to infer latent system states. In this case, the motion dynamics of a driving car is modelled by a linear-like dynamical model, with DNNs to extract useful features from visual observations, and SSMs to estimate and predict system states with or even without observations.   
         }
    \label{fig:concept}
\end{figure*}

In this work, we propose \textit{DynaNet} - neural Kalman dynamical models, combining the respective advantages of deep neural networks (powerful feature representation) and state space models (explicit modelling of physical processes).
As illustrated in Figure \ref{fig:concept}, our proposed end-to-end model enables automatic feature extraction from raw data and constructs a time-varying transition model. 
The recursive Kalman Filter is incorporated to provide optimal filtering on the feature state space and estimate process covariance to reason about system behaviour and stability. This allows for accurate system identification in an end-to-end manner without human effort. 

Our DynaNet can learn a linear-like state-space-model directly from raw data. This linear-like structure is a drop-in replacement for the typical recurrent neural network estimator. {Specifically, our contributions are three-fold:

(1) We propose a novel hybrid model with differentiable Kalman filter that is adopted on the feature level instead of system states level for latent state inference. 

(2) We design a strategy to ensure the stability of learned dynamical model by resampling the transition matrix from Dirichlet distribution, in which the system proves to be stable with a probability of one.

(3) With the design of neural emission model to connect observations with full states, our DynaNet can cope with  a number of challenging situations e.g. when only partial/corrupted observations are available or even without any observations. 
}

To demonstrate the effectiveness of the proposed technique, we conducted extensive experiments and a systematic study on challenging real world motion estimation and prediction tasks, including visual odometry, visual-inertial odometry and motion prediction. We show how the proposed method outperforms the state-of-the-art deep-learning techniques, whilst yielding valuable interpretable information about model performance. {The interpretability analysis discovers the interesting relation between sensor data quality and the explicit model terms}.

The rest of this paper is organized as follows: Section II reviews the relevant work; Section III presents our proposed neural Kalman dynamical model; Section IV evaluates our proposed model applied to three different tasks, i.e. visual egomotion estimation and prediction, visual-inertial navigation, and motion prediction through extensive experiments; Section V finally discusses conclusions.

\section{Related Work}
\subsection{State-Space-Models (SSMs)}
State-space-model is a convenient and compact way to represent and predict system dynamics. 
In classical control engineering, system identification techniques are widely employed to build parametric SSMs \cite{Dudley1979, yu2017incremental}. In order to alleviate the effort of analytic description, data-driven methods, such as Gaussian Processes \cite{Kocijan2005}, or Expectation–Maximization (EM) \cite{Ghahramani1999}, emerged as alternatives to identify nonlinear models. Linear dynamic model, e.g. Kalman filtering has been explored to combine with recurrent neural network that ensures the convergence of neural network training \cite{wang2011convergence}. 
With advances in deep neural networks (DNNs), deep SSMs have been recently studied to handle very complex nonlinearity. Specifically,  Backprop KF \cite{Haarnoja2016} and DPF \cite{Jonschkowski2018} used DNNs to extract effective features and feed them to a predefined physical model (i.e. conditioned on algorithmic priors) to improve filtering algorithms. \cite{karkus2018particle} incorporates particle filter as an algorithmic prior into the neural network for visual localization. Besides feature extraction, DNNs have also been used in re-parameterizing the transition matrix in SSMs from raw data \cite{Krishnan2017,Fraccaro2016,Fraccaro2017,Karl2016}. 
Unlike prior art, our work exploits recent findings on stable dynamical models \cite{Umlauft2017} and uses resampling to generate a transition matrix from the Dirichlet distribution, whose concentration is learnt via a neural network. The specific Dirichlet distribution ensures the stability of dynamic systems, which is an important yet absent property of previous DNNs based SSMs. 

  \begin{figure*}
     	\centering
         \includegraphics[width=0.85\textwidth]{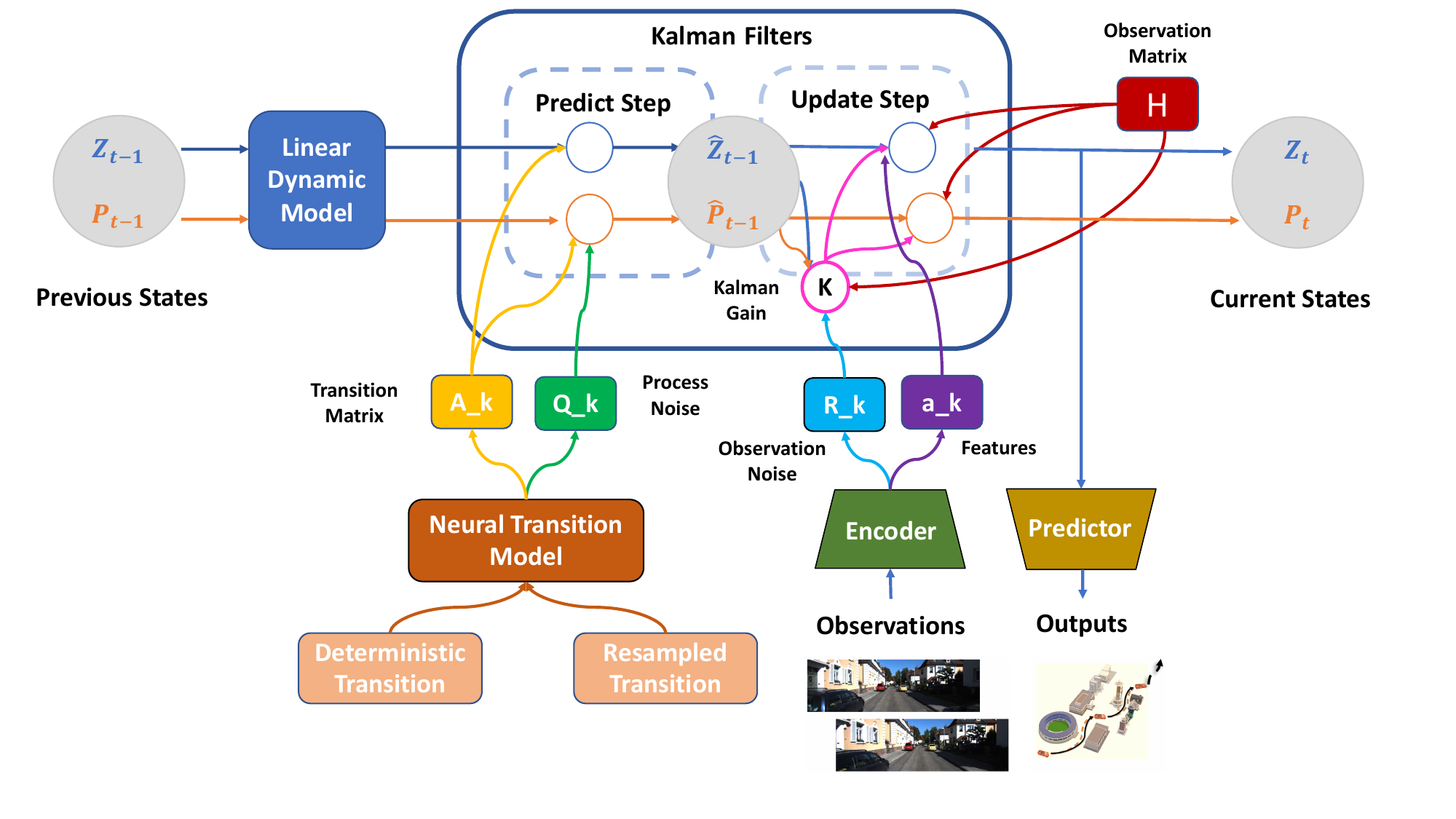}
         \caption{The DynaNet framework consists of the neural observation model to extract latent states $\mathbf{z}$, the neural transition model to generate the evolving relation $\mathbf{A}$, and a recursive Kalman Filter to infer and predict system states. $\mathbf{P}$ is the covariance matrix of latent states, $\mathbf{Q}$ and $\mathbf{R}$ are process noise matrix and observation noise matrix respectively.
         }
         \label{fig:framework}
     \end{figure*}

\subsection{Motion Estimation}
Motion estimation has been studied for decades and plays a central role in robotics and autonomous driving. Conventional visual odometry/SLAM methods rely on multiple-view geometry to estimate motion displacement between images \cite{Nister2004,Davison2007,Newcombe2011,Engel2013,Engel2014,Forster2014,Montiel2015}. Owning to the huge availability and complementary property of inertial and visual sensors, integrating these two sensor modalities has raised increasing attentions to give more robust and accurate motion estimates \cite{Forster2015}. A large portion of work in this direction is visual-inertial odometry, where filtering \cite{Li2013b,Bloesch2015} and nonlinear optimisation \cite{Leutenegger2015,Forster2015,Qin2018} are two mainstream model-based methods for sensor fusion. Meanwhile, recent studies also found that the methods using data-driven DNNs are able to provide competitive robustness and accuracy over some model-based methods. These deep learning methods often use convolutional neural networks (ConvNets) to discover useful geometry features from images for effective odometry estimation \cite{Zhou2017,Zhan2018,Yin2018,Tang2019,kashyap2020sparse}, and/or employ RNNs to model the temporary dependency in motion dynamics \cite{Clark2017a,Mirowski2017,Wang2017,Henriques2018}.
{Besides self-motion estimation in robotics and autonomous driving, RNNs have also been introduced to model human dynamics, and address the problem of human-skeleton motion prediction \cite{fragkiadaki2015recurrent,martinez2017human}.
To improve the capacity of long-term prediction, \cite{tang2018long} leverages temporal attention mechanism to predict next step motion based on all historical information. Furthermore,
\cite{shu2021spatiotemporal} considers both the spatial and temporal relations of human skeleton motions, and proposes a novel skeleton-joint attention with RNNs to achieve better performance in the task of human motion prediction.
} 
Nevertheless, DNN based methods are hard to interpret or expect/modulate their behaviours \cite{Sunderhauf2018}. Motivated by this, our DynaNet aims to bridge the gap of performance and interpretability through a deeply coupled framework of model-based and DNN-based methods. 

\section{Neural Kalman Dynamic Models}
We consider a time-dependent dynamical system, governed by a complex evolving function $f$:
    \begin{equation}
        \label{eq: dynamic system}
        \mathbf{z}_{t} = f(\mathbf{z}_{t-1}, \mathbf{w}_t)
    \end{equation}
where $\mathbf{z} \in \mathbb{R}^d$ is $d$-dimensional latent state, $t$ is the current timestep, and $\mathbf{w}$ is a random variable capturing system and measurement noise. 
The evolving function $f$ is assumed to be Markovian, describing the state-dependent relation between latent states $\mathbf{z}_t$ and $\mathbf{z}_{t-1}$.
The model in Equation (\ref{eq: dynamic system}) can be reformulated as a linear-like structure, i.e. the state-dependent coefficient (SDC) form, with a time-varying transition matrix $\mathbf{A}$:
    \begin{equation}
        \label{eq:linear structure}
        \mathbf{z}_t = \mathbf{A}_t \mathbf{z}_{t-1}
    \end{equation}
Notably, the system nonlinearity is not restricted by this linear-like structure, as there always exists a SDC form $f(z) = A(z)z$ to express any continuous differentiable function $f$ with $f(0) = 0$ \cite{Cimen2008}.

In this regard, our problem of the dynamic model is \emph{how to recover the latent states $\mathbf{z}$ and their time-varying transition relation $\mathbf{A}$ from high-dimensional measurements $\mathbf{x}$ (e.g. a sequence of images), without resorting to a hand-crafted physical model}.

This work aims to construct and reparameterize this dynamic model using the expressive power of deep neural networks and explicit state-space models.
Figure \ref{fig:framework} shows the main framework, which will be discussed in details in the following subsections.
To avoid confusion, in the rest of this paper, \emph{latent} states $\mathbf{z}$ are exclusively used for dynamical models and \emph{hidden} states $\mathbf{h}$ exclusively represent the neurons in a deep neural network. In the meantime, we will use sensor measurements and observations interchangeably.

   \begin{figure*}
     	\centering
         \includegraphics[width=0.7\textwidth]{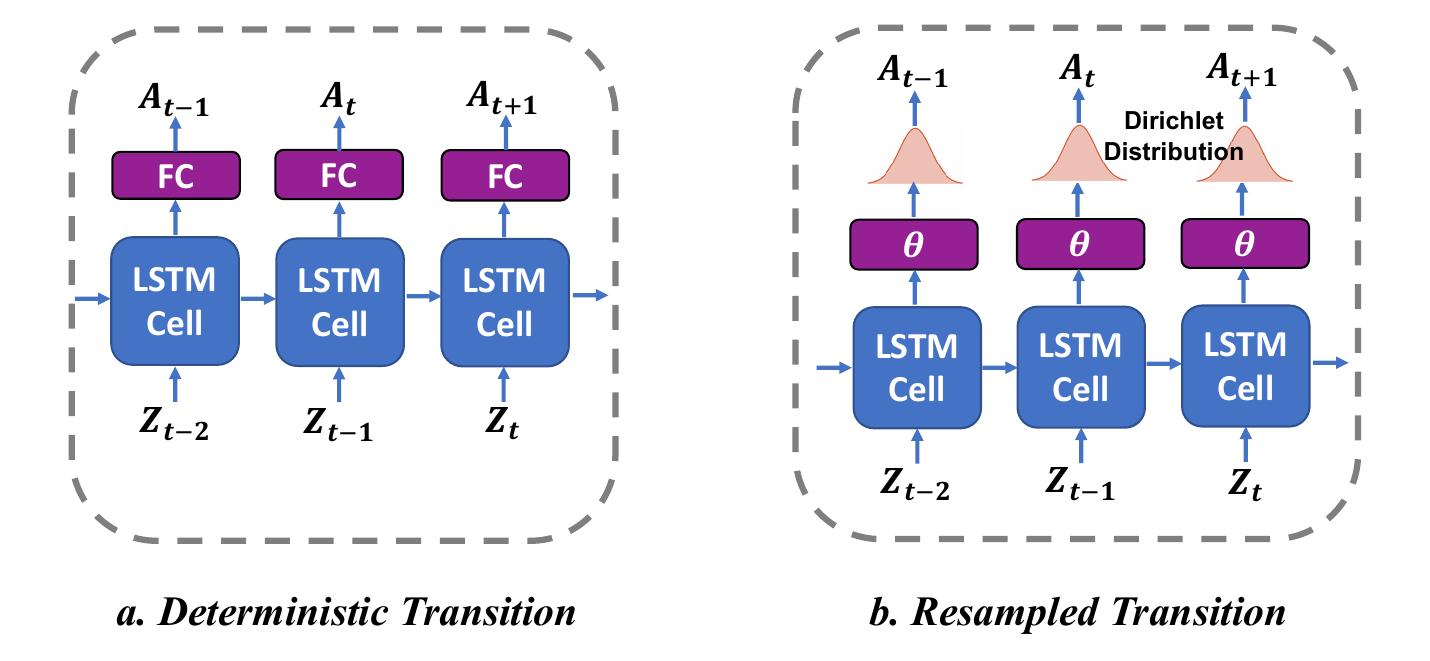}
         \caption{Two transition generation strategies - (a) deterministic transition or (b) resampled transition are proposed to achieve desired system behaviours. The transition matrix is generated by a recurrent neural network, i.e. LSTM in this work conditioned on previous latent system states.
         }
         \label{fig: transition}
     \end{figure*}

\subsection{Neural Emission Model}

Intuitively, the system states containing useful information often lie in a latent space that is different from the original measurements. 
For example, given a sequence of images (the sensor measurements), the key system states of visual odometry are velocity, orientation and position. Nevertheless, it is non-trivial for conventional models to formulate a temporal linear model that can precisely describe the relation between these physical representations.

Rather than explicitly specifying physical states as in a classical SSM, we use a deep neural network to automatically extract the latent state features whilst forcing them to follow the linear-like relation in Equation (\ref{eq:linear structure}). This can be achieved automatically by optimizing the model via stochastic gradient descent and backpropagation algorithms. This linearization is particularly useful as it allows us to directly use a Kalman Filter for state feature inference.
Note that the differentiable Kalman filtering in our DynaNet model performs on the high dimensional latent feature space rather than the physical state space (e.g. velocity, orientation and position) as in Backprop KF \cite{Haarnoja2016} and DPF \cite{Jonschkowski2018}.

In our neural emission model, an encoder $f_\text{encoder}$ is used to extract both features $\mathbf{a}_t$ and an estimation of uncertainty $\bm{\sigma}^a_t$ from the observations $\mathbf{x}_t$ at timestep $t$: 
    \begin{equation}
        \mathbf{a}_t, \bm{\sigma}_t = f_\text{encoder}(\mathbf{x}_t).
    \end{equation}
The features $\mathbf{a}$ act as observations of the latent feature state space. The coupled uncertainties $\bm{\sigma}$ represent the measurement belief that is transformed into the observation noise $\mathbf{R}$ in a Kalman Filter. {$\mathbf{a}$ and $\bm{\sigma}$ are further used in the update stage of a differentiable Kalman filter in Equation (14), which forces them to follow the distribution of KF parameters during end-to-end optimization. Thus, unlike VAE\cite{kingma2013auto,Karl2016}, the two terms are not directly and explicitly constrained in Equation (3) to follow a prior distribution, but leave the learning model to search for suitable latent states and uncertainty estimation that can construct a linear-like structure and exploit the full capacity of differentiable Kalman filter.}
However, the observations $\mathbf{a}$ are sometimes unable to provide sufficient information for all latent states $\mathbf{z}$ in a dynamical system, for example, the occasional absences of sensory data. Hence, a deterministic emission matrix $\mathbf{H}$ is defined to connect with the full latent states $\mathbf{z}$:
    \begin{equation}
        \mathbf{a}_t = \mathbf{H} \mathbf{z}_t 
    \end{equation}
In a practical setting, when the extracted features contain all the information for dynamical systems, the emission matrix $\mathbf{H}$ is set to d-dimensional identity matrix $\mathbf{I}_d$ as features and states are identical at this moment. On the other side, the identity matrix needs to adapt to $\mathbf{H}= [\mathbf{I}_m, \mathbf{0}_{m\times(d-m)}]$, when observations only give rise to $m$ features. In this case, the rest $(d-m)$ latent states will be attained from historical states. Our experiment in Section \ref{sub:visual_inertial_navigation} demonstrates the superiority this neural emission model in addressing the issue of observation absences for sensor fusion in visual-inertial odometry.  

\subsection{Neural Transition Model}
In a SSM, the temporal evolution of latent states is determined by the transition matrix $\mathbf{A}$ as in Equation \ref{eq:linear structure}. Obviously, the transition matrix is of considerable importance as it directly describes the governing mechanism of a system. Nevertheless, such a matrix is difficult to specify manually, especially when it is time-varying.
Figure \ref{fig: transition} demonstrates two methods we propose to estimate $\mathbf{A}$ on the fly, based on prior system states: (a) a deterministic way to learn it end-to-end from raw data; (b) a stochastic way to resample it from a distribution, e.g. the Dirichlet distribution in this work.  {The deterministic transition and resampled transition are two individual strategies to generate transition matrix. The parameters of LSTM in these two modules are separately learned from data.}
We will explain them accordingly in what follows.

\subsubsection{Deterministic Transition}

Intuitively, a movement change depends on historical system states, which are encoded in the latent states $\mathbf{z}_{0:t-1}$. Prior works mostly apply a recurrent neural network (RNN) to specify dynamic weights for choosing and interpolating between a fixed number of different transition modes \cite{Karl2016,Fraccaro2017}. Inspired by \cite{Rangapuram2018}, our model generates the transition matrix $\mathbf{A}$ directly from the history of latent states $\mathbf{z}$. 

In this deterministic transition model, the dependence of the transition matrix on historical latent states is specified by a RNN. This RNN recursively processes previous hidden states $(\mathbf{z}_{t-1}, \mathbf{h}_{t-1})$ of the dynamic model and Long Short-Term Memory (LSTM) \cite{hochreiter1997long,greff2016lstm} cell respectively, and outputs the time-dependent transition matrix $\mathbf{A}_t$ via:
    \begin{equation}
        \mathbf{A}_t = \text{LSTM} (\mathbf{z}_{t-1}, \mathbf{h}_{t-1}) ,
    \end{equation} 
where $\mathbf{z}_{t-1}$ is the latent states of dynamical system at the timestep $(t-1)$, $\mathbf{h}_{t-1}$ is the hidden states of LSTM module, $\mathbf{A}_t$ is the learned transition matrix at current timestep.
    
\subsubsection{Resampled Transition}
Powerful deep neural networks are able to approximate dynamical models effectively from data. However, these learned `black-box' models are difficult to be interpreted or modulated. Especially, in safety-critical applications (e.g. self-driving vehicles), the lack of model behavioural indicator and the absence of system reliability largely limit the adoption of these learning models. The stability of a dynamical system is essential and fundamental to autonomous systems, as it guarantees that the predictions of dynamic models will not change abruptly, given slightly perturbed inputs.
Unfortunately, this desirable property is not ensured by most pure DNN models.
Therefore, we here aim to ensure the stability of learned dynamical model by resampling the transition matrix from a specific distribution.

The linear-like SSM structure in Equation (\ref{eq:linear structure}) allows for a quadratic Lyapunov stability analysis, whilst advances in stochastic optimisation allow to construct neural probabilistic models \cite{Schulman2015}. 
We thus propose to resample the transition matrix from a predefined probability distribution to enforce the desired stability. Based on the findings in \cite{Umlauft2017}, if the state transition follows a Dirichlet distribution in a positive system, it will lead to a model being asymptotically stable i.e. it will be Bounded Input Bounded Output (BIBO) stable. Constructing the stochastic variable and determining the parameters of the distribution are easy to achieve with the widely used reparameterisation trick \cite{Jankowiak2018}.  

To further reduce hand-crafted engineering, the concentration $\bm{\alpha}$ of the Dirichlet distribution in our framework is generated from historical system states via an LSTM based RNN:
    \begin{equation}
        \bm{\alpha} = \text{LSTM} (\mathbf{z}_{t-1}, \mathbf{h}_{t-1}) ,
    \end{equation}
where $\mathbf{z}_{t-1}$ is the latent states of dynamical system at the timestep $(t-1)$, $\mathbf{h}_{t-1}$ is the hidden states of LSTM module.
A small Gaussian random noise is also added in this process to improve model robustness. At each timestep, a realisation of the transition matrix $\mathbf{A}$ is drawn from the constructed Dirichlet distribution:
    \begin{equation}
        \mathbf{A}_t \sim \text{Dirichlet} (\bm{\alpha}) 
    \end{equation}
Note that in DynaNet the transition states are in the latent feature space rather than the final target states (e.g. the states of orientation and position in VO). The latent features are extracted by the encoder, which ensures the transition states strictly positive through a ReLU activation and a tiny random positive number on the last layer of the DNN based encoder. 

Our DynaNet extends the work \cite{Umlauft2017} of nearest neighbor method based stable system to a deep neural network based approach, for modelling more complex nonlinear dynamics. Analytically, the following proof supports our proposed system:
\textit{given a DNN based dynamical system $\mathbf{z}_{t+1}=\mathbf{A}(\bm{\alpha})\mathbf{z}_t$,
where $\mathbf{z}_t \in \mathbb{R}^{d}$ is $d$-dimensional system state at the timestep $t$, extracted by a neural network from raw data, $\mathbf{A}(\bm{\alpha}) \in \mathbb{R}^{d \times d}$ is the transition matrix generated by a neural network with a concentration $\bm{\alpha}$, if the transition matrix $\mathbf{A}$ is constructed from a Dirichlet distribution$\mathbf{A} \sim \text{Dir}(\theta (\bm{\alpha}))$,
this dynamical system is asymptotically stable}.

\textbf{Proof:}
By resampling the transition matrix $\mathbf{A} \in \mathbb{R}^{d \times d}$ from a Dirichlet Distribution, the elements inside $\mathbf{A}$ satisfy:
\begin{equation}
    \begin{split}
        &\mathbf{A}_{(i,j)} > 0, \quad \mathbf{A}_{(i,j)} < 1 \quad \forall i,j = 1...d, \quad \\
        &\text{and} \; \sum_{i=1}^d \sum_{j=1}^d \mathbf{A}_{(i,j)} = 1
    \end{split}
\end{equation}
We can have
\begin{equation}
    \sum_{j=1}^d \mathbf{A}_{(i,j)} < 1 \; \forall i \implies \max_{i=1:d} \sum_{j=1}^d \mathbf{A}_{(i,j)} < 1
\end{equation}
If all elements inside $\mathbf{A}$ are strictly positive, then the Maximum absolute Row Sum Norm $\| \mathbf{A} \|_{\infty}$ follows: 
\begin{equation}
    \| \mathbf{A} \|_{\infty} = \max_{i=1:d} \sum_{j=1}^d |\mathbf{A}_{(i,j)}| < 1
\end{equation}
The hidden feature states $\mathbf{z}_{k+M}$ at the time step $t+M$ is derived from the system states $\mathbf{z}_k$ at the time step $t$ and $M$ consecutive transition matrix via:
\begin{equation}
    \begin{split}
        \| \mathbf{z}_{k+M} \|_{\infty} = \| \prod_{m=1}^M \mathbf{A}^m \mathbf{z}_k \|_{\infty} 
        &\leq \prod_{m=1}^M \| \mathbf{A}^m \|_{\infty} \|\mathbf{z}\|_{\infty} \\
        &\leq (\max_m \|\mathbf{A}^m\|_{\infty})^M\|\mathbf{z}_k\|_{\infty}
    \end{split}
\end{equation}
As the $M$ is infinite, the system states $\mathbf{z}_{k+M}$ will become:
\begin{equation}
    \| \mathbf{z}_{k+M} \|_{\infty} \xrightarrow{M \rightarrow \infty} 0
\end{equation}
Therefore, the system is stable with a probability of one.

\subsection{Prediction and Inference with a Kalman Filter} 
\label{sub:prediction_and_inference_via_kalman_filter}
The neural emission model estimates system states from noisy sensor measurements, while the generated transition model describes the system evolution and predicts the system states with previous ones. However, uncertainties exist in both of them and motivate us to integrate a Kalman Filter into our framework. The Kalman Filter recursively deals with the uncertainties, and produces a weighted average of the state predictions and fresh observations. With the aforementioned neural emission and transition models, the prediction and inference are performed on the feature state space and follow a standard Kalman filtering pipeline. We also note that the Kalman Filter's gain controls how much to update the residual error (i.e., the difference between prediction and observation), which is a useful metric to represent the relative quality of measurements (as shown in Section~\ref{sub:kalman_gain}). 

More specifically, the Kalman Filter consists of two blocks: prediction and update. In the prediction stage, prior estimates of the mean value and covariance $(\mathbf{z}_{t|t-1}, \mathbf{P}_{t|t-1})$ at the current timestep are derived from the posterior state estimates $(\mathbf{z}_{t-1|t-1}, \mathbf{P}_{t-1|t-1})$ in the previous timestep:
    \begin{align}
    \begin{split}
        \mathbf{z}_{t|t-1} &= \mathbf{A}_t \mathbf{z}_{t-1|t-1},
    \\
        \mathbf{P}_{t|t-1} &= \mathbf{A}_t \mathbf{P}_{t-1|t-1} \mathbf{A}_t^T + \mathbf{Q}_t .
    \end{split}
    \end{align}
When current observations $\mathbf{a}_t$ are available, the update process allows us to produce a posterior mean and covariance of hidden states $(\mathbf{z}_{t|t}, \mathbf{P}_{t|t})$ as follows:
    \begin{align}
    \label{eq: kalman update}
    \begin{split}
        \mathbf{r}_t &= \mathbf{a}_t - \mathbf{H}_t \mathbf{z}_{t|t-1} ,
    \\
        \mathbf{S}_t &= \mathbf{R}_t + \mathbf{H}_t \mathbf{P}_{t|t-1} \mathbf{H}_t^T ,
    \\
        \mathbf{K}_t &= \mathbf{P}_{t|t-1} \mathbf{H}_t^T \mathbf{S}_t^{-1} ,
    \\
        \mathbf{z}_{t|t} &= \mathbf{z}_{t|t-1} + \mathbf{K}_t \mathbf{r}_t ,
    \\
        \mathbf{P}_{t|t} &= (\mathbf{I} - \mathbf{K}_t \mathbf{H}_t) \mathbf{P}_{t|t-1} ,
    \end{split}
    \end{align}
where $\mathbf{r}$ is the residual error (aka. innovation), $\mathbf{S}$ is the residual covariance and $\mathbf{K}$ is Kalman gain. In contrast to hand-tuning process noise $\mathbf{Q}$ and measurement noise $\mathbf{R}$ in a conventional KF, these two terms are jointly learned by our proposed neural dynamical model. Finally, the predictor (e.g. a FC network) outputs the target values $\mathbf{y}_t$ from the estimated optimal hidden states $\mathbf{z}_{t|t}$:
    \begin{equation}
        \mathbf{\tilde{y}}_t = f_{\text{predictor}} (\mathbf{z}_{t|t}) .
    \end{equation}
In the case that current measurements i.e. $\mathbf{a}_t$ are unavailable, the reconstructed values $\mathbf{\hat{y}}_t$ are inferred from the prior estimate $\mathbf{z}_{t|t-1}$:
    \begin{equation}
        \mathbf{\hat{y}}_t = f_{\text{predictor}} (\mathbf{z}_{t|t-1}) .
    \end{equation}
All parameters $\theta$ in our model are end-to-end learned with a mean square loss function. This loss function jointly compares the ground truth $\mathbf{y}_t$ with posterior predictions $\mathbf{\tilde{y}}_t$ and prior prediction $\mathbf{\hat{y}}_t$:
    \begin{equation}
        L(\theta) = \frac{1}{T} \displaystyle\sum_{t=1}^{T} (|| \mathbf{y}_t - \mathbf{\tilde{y}}_t ||^2 + || \mathbf{y}_t - \mathbf{\hat{y}}_t ||^2)
    \end{equation}
    
     \begin{table*}[t]
  		\caption{The performance of visual odometry on the KITTI odometry dataset for motion estimation, reported in the RMSE of translation (\%) and orientation ($^{\circ}$)} 
  		\label{tab: vo}
  		\centering
  		\small
  		\begin{tabular}{c|c|c|c|c||c|c|c|c}
    	& {ORB-SLAM} & VISO2 & SfmLearner & Bian et al. & DeepVO (LSTM) & Ours (Deter.) & Ours (Dirichlet) \\
    		 \hline
    		09 & {45.52\%, 3.10$^{\circ}$} &  18.06\%, 1.25$^{\circ}$  & 17.84\%, 6.78$^{\circ}$ & 11.2\%, 3.35$^{\circ}$ &  8.01\%, 3.10$^{\circ}$ & 6.43\%, 2.19$^{\circ}$ & \textbf{4.97\%}, \textbf{2.10$^{\circ}$}\\
    		10 & {\textbf{6.39\%}, 3.20$^{\circ}$} & 26.10\%, 3.26$^{\circ}$ & 37.91\%, 15.78$^{\circ}$ & 10.1\%, 4.96$^{\circ}$ &  8.53\%, 2.41$^{\circ}$ & 8.35\%, 2.39$^{\circ}$ & 9.08\%, \textbf{2.15$^{\circ}$} \\
    		\hline
    		ave & {25.95\%, 3.15$^{\circ}$} & 22.08\%, 2.25$^{\circ}$ & 27.87\%, 11.28$^{\circ}$ & 10.65\%, 4.15$^{\circ}$ & 8.27\%, 2.75$^{\circ}$ & 7.39\%, 2.29$^{\circ}$ & \textbf{7.03\%}, \textbf{2.12$^{\circ}$}
  		\end{tabular}
  		\begin{itemize}
              \footnotesize{
                \item $t_{rel}(\%)$ is the average translational RMSE drift (\%) on lengths of 100m-800m.
                \item $r_{rel}(^{\circ})$ is the average rotational RMSE drift ($^{\circ}$/100m) on lengths of 100m-800m.
                \item The DeepVO (LSTM), our proposed deterministic (Ours (Deter.)) and Dirichlet (Ours (Dirichlet)) model are trained on Sequence 00 - 08 of the KITTI dataset \cite{Geiger2013} with same hyperparameters for a fair comparison, and tested on Sequence 09 and 10.
            }
        \end{itemize}
	\end{table*}

\section{Experiments}
We systematically evaluate our system through extensive experiments including (1) Section~\ref{sub:visual_egomotion_estimation} - pose estimation for visual odometry, (2) Section~\ref{sub:visual_inertial_navigation} -  pose estimation for visual-inertial odometry and (3) Section~\ref{sub: motion prediction} - motion prediction without observations or with partial observations. Moreover, an interpretability study is also conducted in Section~\ref{sub:kalman_gain}.

\subsection{Datasets, Baselines and Experiment Setups} 
\label{sub:experimental_setup_and_datasets}
\subsubsection{Datasets}
To evaluate our proposed DynaNet models, we used public datasets to conduct experiments: KITTI Odometry dataset \cite{Geiger2013} for visual pose estimation and prediction, and KITTI Raw Dataset \cite{Geiger2013} for visual-inertial pose estimation and prediction.

\textbf{KITTI Odometry Dataset \cite{Geiger2013}} is a commonly used benchmark dataset that contains 11 sequences (00-10) with images collected by car-mounted cameras and ground-truth trajectories provided by GPS. We used it for visual odometry experiment, with \textit{Sequences 00-08} for training and \textit{Sequences 09, 10} for testing. The images and ground truth are collected at 10 Hz. We chose the sequence length as 5, and thus generated a total of 20373 sub-sequences from training set to train neural models.

\textbf{KITTI Raw Dataset \cite{Geiger2013}} contains both raw images (10 Hz) and high-frequency inertial data (100Hz). Since inertial data are only available in the unsynced data packages, we selected the raw files with the corresponding to KITTI Odometry Dataset. Inertial data and images are manually synchronized according to their timestamps. We adopted the same data split mentioned above, discarding \textit{Sequence 03} as its raw data is unavailable. Thus, in this experiment, we used \textit{Sequences 00, 01, 02, 04, 05, 06, 07, 08} for training and \textit{Sequences 09, 10} for testing. We chose the sequence length as 5, and thus generated 20373 sub-sequences from training set to train neural models. We chose the sequence length as 5, and thus generated a total of 20361 sub-sequences from training set to train deep neural networks.

\subsubsection{Baselines}

In the visual odometry experiment, we compare our DynaNet models with three representative three deep learning based VO models, i.e. SfmLearner \cite{Zhou2017}, Bian et al. \cite{bian2019unsupervised} and DeepVO \cite{Wang2017}.
DeepVO shares the same architecture as in our models including the encoder and predictor, but uses the 2-layers LSTM \cite{greff2016lstm} to estimate system latent states. This can be viewed as an ablation study, and we keep their dimension of hidden states (128) the same as our models for a fair comparison.
Except learning based baselines, our model is also compared with {two} classical monocular VO system, i.e. {ORB-SLAM \cite{Montiel2015}} and VISO2 \cite{geiger2011stereoscan}. 
{ORB-SLAM \cite{Montiel2015} is monocular visual SLAM algorithm based on hand-crafted features and multi-view geometry. Its loop closing module is disabled for a fair comparison with odometry estimation. VISO2 \cite{geiger2011stereoscan} is a monocular VO algorithm, and implemented as an official baseline on KITTI dataset.}

In the visual-inertial odometry (VIO) experiment, we chose a state-of-the-art learning based VIO approach, i.e. VINet \cite{Clark2017a} as our baseline. VINet shares similar structure as in our models, but it uses a two-layers LSTM rather than our DynaNet module. 
{A popular classical model-based VIO system, i.e.  mono-VINS \cite{Qin2018} is also adopted here as baseline.
Mono-VINS \cite{Qin2018} is a tightly-coupled sliding window-based optimization approach for visual inertial odometry, which achieves state-of-the-art performance on several VIO datasets. }

	\begin{figure*}
    	\centering
        \begin{subfigure}[t]{0.4\textwidth}
        	\includegraphics[width=\textwidth]{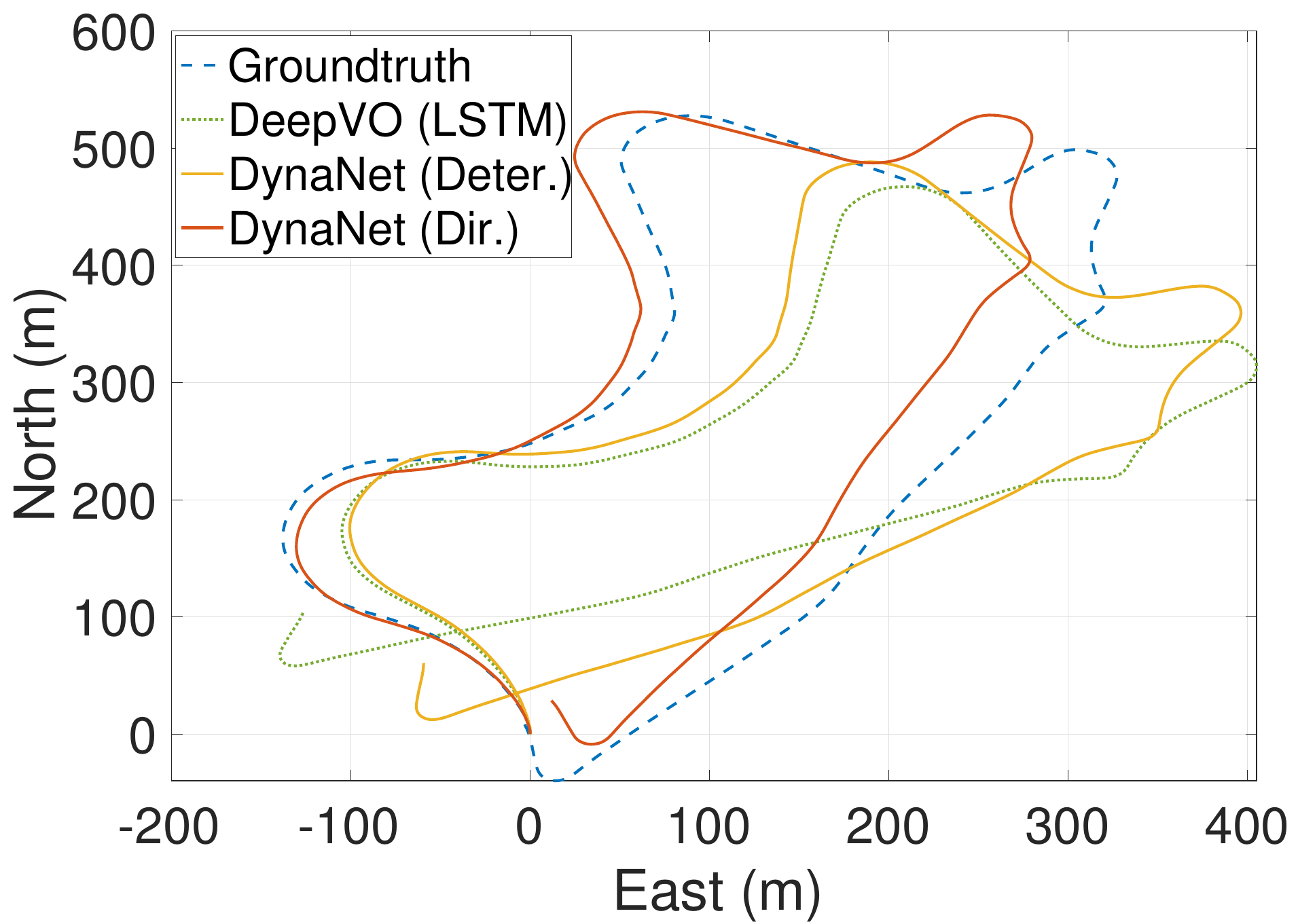}
        	\caption{\label{fig: trajectory 09} Poses estimation in Seq 09}
        \end{subfigure}
        \begin{subfigure}[t]{0.4\textwidth}
        	\includegraphics[width=\textwidth]{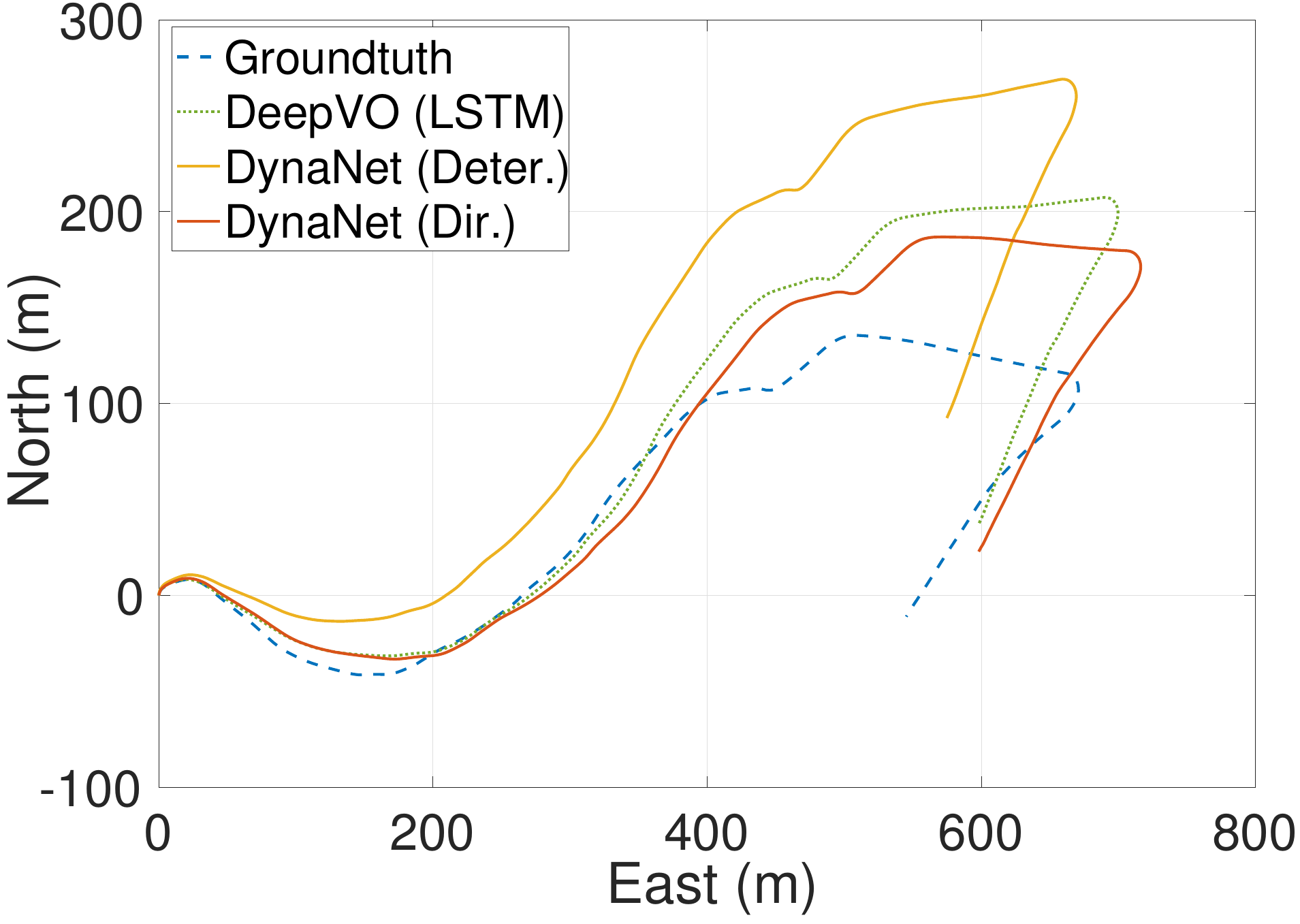}
        	\caption{\label{fig: trajectory 10} Poses estimation in Seq 10}
        \end{subfigure}
        \caption{\label{fig: trajectories vo} The testing trajectories on (a) Sequence 09 and (b) Sequence 10 of the KITTI dataset indicate that our models produce robust and accurate pose estimates in visual odometry.}
    \end{figure*}

In the motion prediction task, we compare our models with LSTM based approaches.
All the other modules for LSTMs including encoder and predictor, and the dimension of hidden states (128) are kept the same as in our proposed models for a fair comparison. 
{Besides, we implemented an attention based LSTM approach to show how attention mechanism improves the motion prediction in future steps. As the approach is an end-to-end model that directly deals with high-dimensional raw images, at each time step the attention mechanism aggregates the features extracted from visual data or visual/inertial data, and then feeds the updated features to LSTM module. The rest modules are kept the same to be compared fairly.}

\subsubsection{Experiment Setups}
We implemented the proposed framework with Pytorch, and trained on a NVIDIA Titan X GPU. The implementation details of DynaNet frameworks can be found at APPENDIX A.
All of our models are trained with the Adam optimizer with a batch-size of 32 and a learning rate of $1e^{-4}$.
The trained models are tested on the test set, in which data have never been seen in the training set.

\subsection{Visual Odometry} 
\label{sub:visual_egomotion_estimation}
Our evaluation starts with a set of visual ego-motion (Visual Odometry) experiments for 6-DoF pose estimation. 
Here, pose estimation means that our model produces 6-DoF pose given sensor data (i.e. images). 
In this experiment, a sequence of raw images are given to models to produce pose transformations, i.e. translation and rotation.

	\begin{table}[t]
  		\caption{The performance of visual-inertial odometry on the KITTI raw dataset for motion estimation, reported in the RMSE of translation (\%) and orientation ($^{\circ}$).} 
  		\label{tab: vio}
  		\centering
  		\small
  		\begin{tabular}{c|c|c|c|c}
    	& {VINS-Mono}  & VINet  & Ours (Deter.) & Ours (Dir.) \\
    		 \hline
    		09 & {41.5\%, 2.41$^{\circ}$} & \textbf{3.89\%}, 2.02$^{\circ}$  & 5.45\%, \textbf{1.24$^{\circ}$} & 4.13\%, 1.39$^{\circ}$ \\
    		10 & {20.3\%, 2.73$^{\circ}$} & 8.99\%, \textbf{1.39$^{\circ}$}  & \textbf{5.49\%}, 2.02$^{\circ}$ & 7.03\%, 2.64$^{\circ}$ \\
    		\hline
    		ave & {30.9\%, 2.57$^{\circ}$} & 6.44\%, 1.70$^{\circ}$ & \textbf{5.47\%}, \textbf{1.63$^{\circ}$}  & 5.58\%, 2.01$^{\circ}$
  		\end{tabular}
  		\begin{itemize}
              \footnotesize{
                \item $t_{rel}(\%)$ is the average translational RMSE drift (\%) on lengths of 100m-800m.
                \item $r_{rel}(^{\circ})$ is the average rotational RMSE drift ($^{\circ}$/100m) on lengths of 100m-800m.
                \item The VINet (LSTM), our proposed deterministic (Ours (Deter.)) and Dirichlet (Ours (Dirichlet)) model are trained on Sequence 00, 01, 02, 04, 05, 07, 08 of the KITTI raw dataset \cite{Geiger2013} with same hyperparameters for a fair comparison, and tested on Sequence 09 and 10.
            }
        \end{itemize}
	\end{table}

Table \ref{tab: vo} reports the performance of our proposed DynaNet models, comparing with other learning based approaches and classical VO algorithm. 
All neural networks were trained above KITTI Odometry dataset with Sequences 00 - 08, while tested with two new sequences (Sequence 09 and 10).
The motion transformations from models are integrated into global trajectories, and them we evaluated them according to the official KITTI metrics, commonly adopted to evaluate VO algorithms, which calculates the average Root Mean Square Errors (RMSEs) of the translation and rotation for all the subsequences in the lengths 100, 200, ..., 800 meters. This evaluation metrics can capture both the global and local drifts of VO systems.

		\begin{table*}[h]
  		\caption{Visual-inertial navigation on the KITTI raw dataset for motion prediction (Translation RMSE [0.01 m]).} 
  		\label{tab: vio prediction}
  		\centering
  		\small
  		\begin{tabular}{c|c|c|c}
    		  & Prediction w/o Inputs & Visual Only Prediction & Inertial Only Prediction \\
    		\hline
    		LSTM (1-layer) & 32.5 & 7.86 & 23.7 \\
    		LSTM (2-layers) & 21.2 & 6.90  & 24.3 \\
    		{LSTM + Attention} & {17.4} & {7.54} & {23.1}   \\
            Ours (Deterministic) & 13.0  & \textbf{6.27}  & \textbf{11.3} \\
            Ours (Dirichlet) & \textbf{12.3} & 6.40  & 12.3 \\
  		\end{tabular}
	\end{table*}
	
	   \begin{table}[t]
  		\caption{Visual odometry on the KITTI odometry dataset for motion prediction (Translation RMSE [0.01 m]).} 
  		\label{tab: motion prediction}
  		\centering
  		\small
  		\begin{tabular}{c|c|c}
    		 & 5 Steps Prediction & 10 Steps Prediction \\
    		 \hline
    		LSTM (1-layer) & 16.8 & 23.4 \\
    		LSTM (2-layers) & 11.0 & 17.7 \\
    		{LSTM + Attention} & {12.1}  & {17.4} \\
            Ours (Deterministic)& 10.8  & 16.3 \\
            Ours (Dirichlet)& \textbf{8.69}  & \textbf{13.5} \\
  		\end{tabular}
	\end{table}

As shown in Table \ref{tab: vo}, our proposed models clearly outperform the baselines of {ORB-SLAM \cite{Montiel2015}}, VISO2 \cite{geiger2011stereoscan} , SfmLearner \cite{Zhou2017}, Bian et al. \cite{bian2019unsupervised} and DeepVO \cite{Zhou2017}, and the largest gain is achieved by our Dirichlet model. Note that the only different between our models and DeepVO is the state estimation part, and we keep the model hyberparameters, e.g. the dimension of hidden states, the same for a fair comparison.
By replacing the LSTM module in DeepVO with our proposed DynaNet, our Deterministic model improves the performance of DeepVO around 10.64\% in translation and 16.73\% in orientation, and our Dirichlet model further improves DeepVO around 14.99\% in translation and 22.91\% in orientation.
This indicates that incorporating the physical prior into neural network benefits learning state estimation from data. And it also implies that the nonlinearities of VO systems are not lost despite the linear-like structures inside our models. 

Figure \ref{fig: trajectories vo} illustrates the trajectories of Sequence 09 and 10 predicted by our models. 
Sequence 09 and 10 are difficult scenarios, as the driving car experienced large movement in height. Our DynaNet models, especially the Dirichlet model, are still capable of providing robust results, which are closer to the groundtruth trajectories, and consistently show competitive performance over LSTM based DeepVO.

\begin{figure}
     	\centering
         \includegraphics[width=0.35\textwidth]{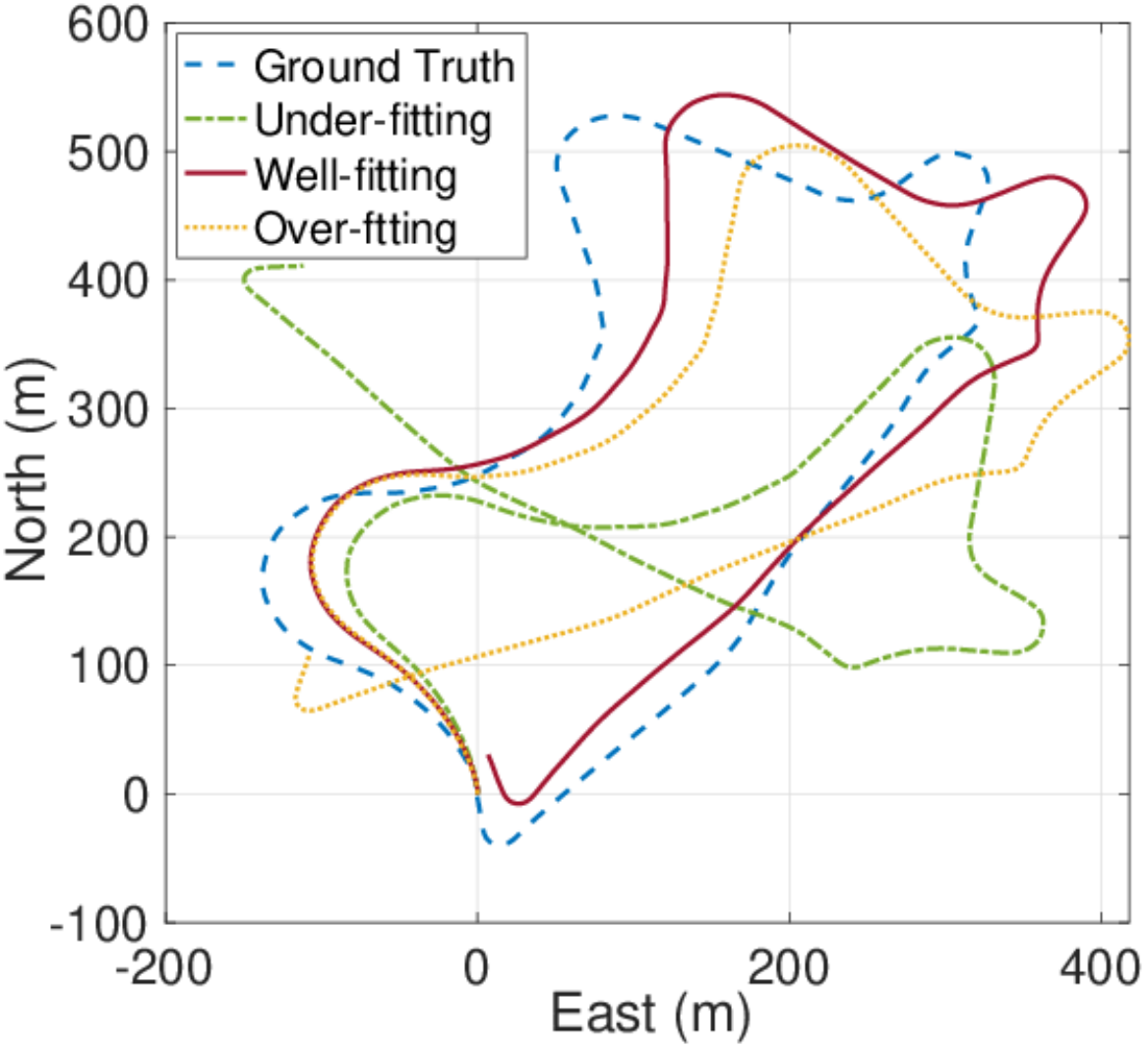}
         \caption{The testing trajectories of our proposed DynaNet model (Dirichlet) on Sequence 9 of the KITTI dataset in the under-fitting, well-fitting and over-fitting conditions.
         }
         \label{fig: training}
     \end{figure}

It must be pointing out that the performance of learning model depends on its training process. Both under-fitting and over-fitting should be avoided for machine learning methods, or their testing performance will be degraded in such circumstances. Figure \ref{fig: training} shows the trajectories of our proposed DynaNet model with Dirichlet distribution on Sequence 9 of the KITTI dataset in three different training stages. Our model is trained for a total of 100 epochs.
When DynaNet model is trained under the well-fitting condition (Epoch 89), the testing trajectory is comparable closer to the ground-truth. However, when our DynaNet is in the under-fitting (Epoch 50) or over-fitting (Epoch 90) condition, it sees larger drifts in the corresponding trajectory.

\subsection{Visual-Inertial Odometry} 
\label{sub:visual_inertial_navigation}
How to effectively integrate and fuse two modalities to provide accurate and robust pose remains a challenging problem. In this experiment, we demonstrate that our proposed models can learn a compact state-space-model for sensor fusion from two modalities, i.e. visual and inertial data. We also show that our DynaNet models enable robust prediction under the circumstances with partial observations in Section \ref{sub: motion prediction}. When the training model is underfitting or overfitting,

\subsubsection{Hyper-parameters Setup}
Initially, a visual encoder and an inertial encoder extract $m$-dimensional visual features $\mathbf{a}_{\text{visual}} \in \mathbb{R}^m$ and $n$-dimensional inertial features $\mathbf{a}_{\text{inertial}} \in \mathbb{R}^n$ separately. These two features are then concatenated together as $\mathbf{a} = [\mathbf{a}_{\text{visual}}, \mathbf{a}_\text{{inertial}}] \in \mathbb{R}^{m+n}$. Notably, our emission matrix is defined as identity matrix $\mathbf{H} = \mathbf{I}_{m+n}$, when both modalities are available. If visual or inertial cues are absent, the emission matrix is changed to $\mathbf{H} = [\mathbf{I}_m, \mathbf{0}_{m \times n}]$ or $\mathbf{H} = [\mathbf{0}_{n \times m}, \mathbf{I}_n]$. The training and testing of the visual-inertial dynamic model follows the same procedures as in visual odometry.

    \begin{figure*}
    	\centering
    	\begin{subfigure}[t]{0.32\textwidth}
    	\includegraphics[width=\textwidth]{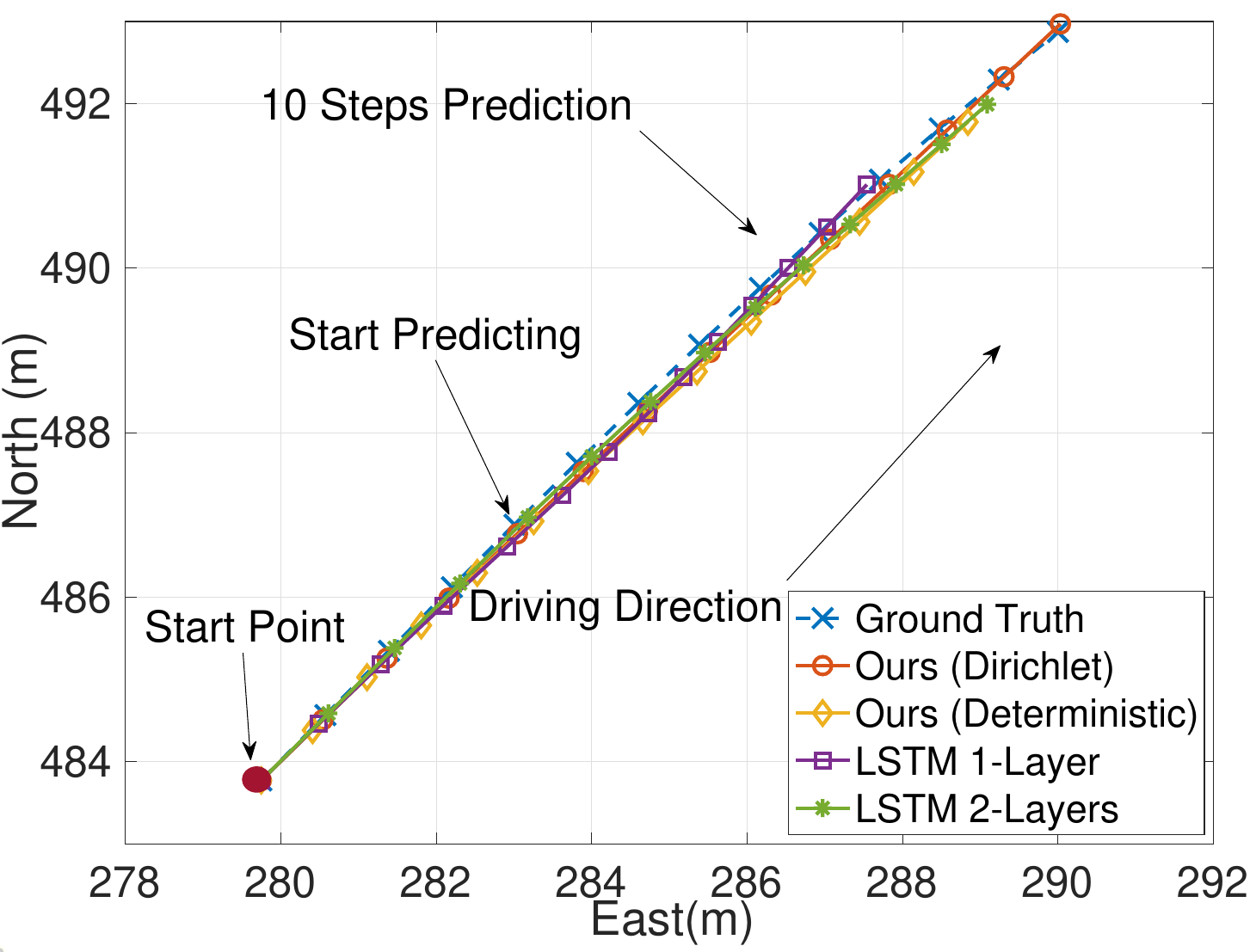}
        	\caption{\label{fig: line best} Line prediction in the good case}
        \end{subfigure}
         \begin{subfigure}[t]{0.32\textwidth}
    	\includegraphics[width=\textwidth]{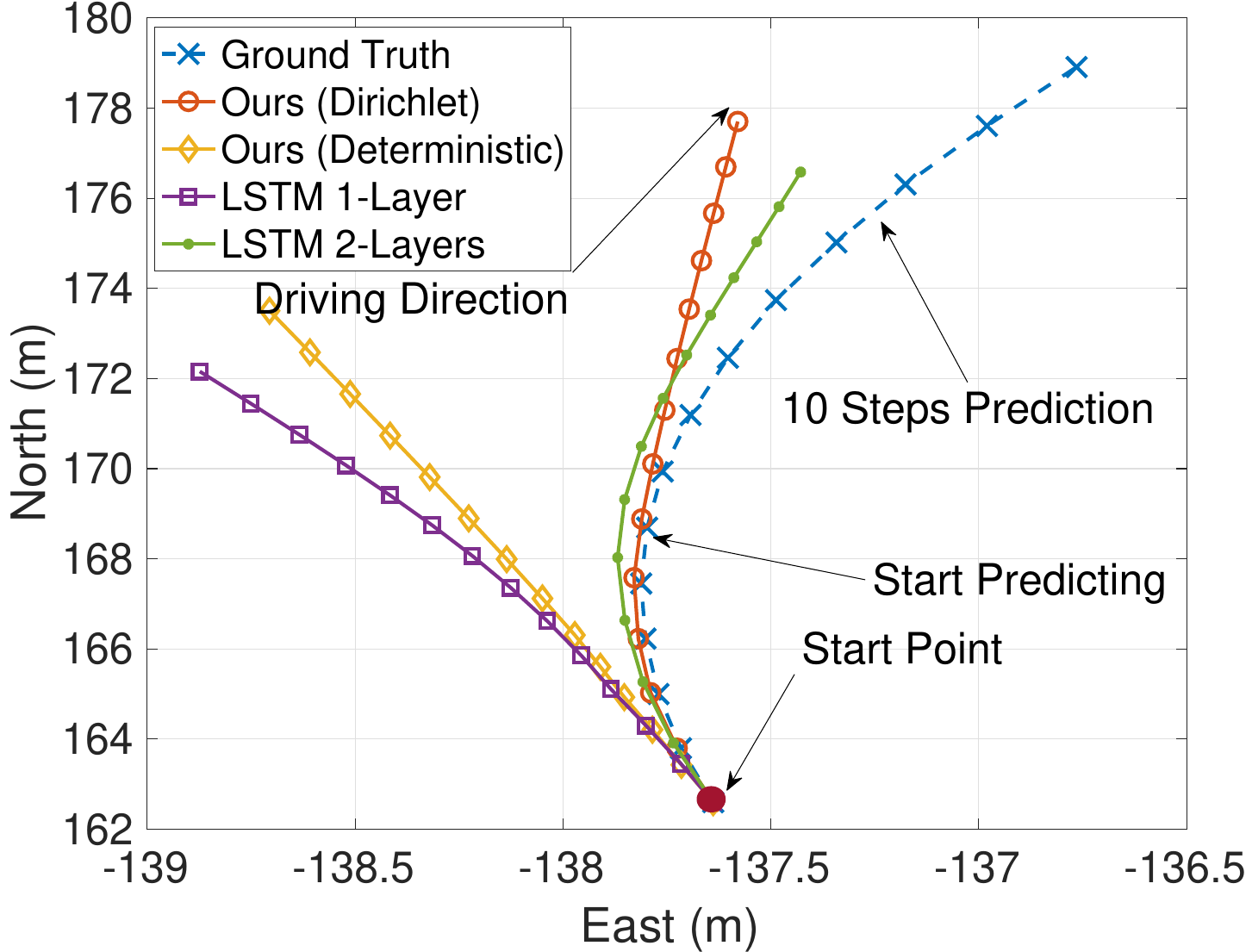}
        	\caption{\label{fig: turning good} Turning pred. w/o inertial data in the good case}
        \end{subfigure}
         \begin{subfigure}[t]{0.32\textwidth}
        \includegraphics[width=\textwidth]{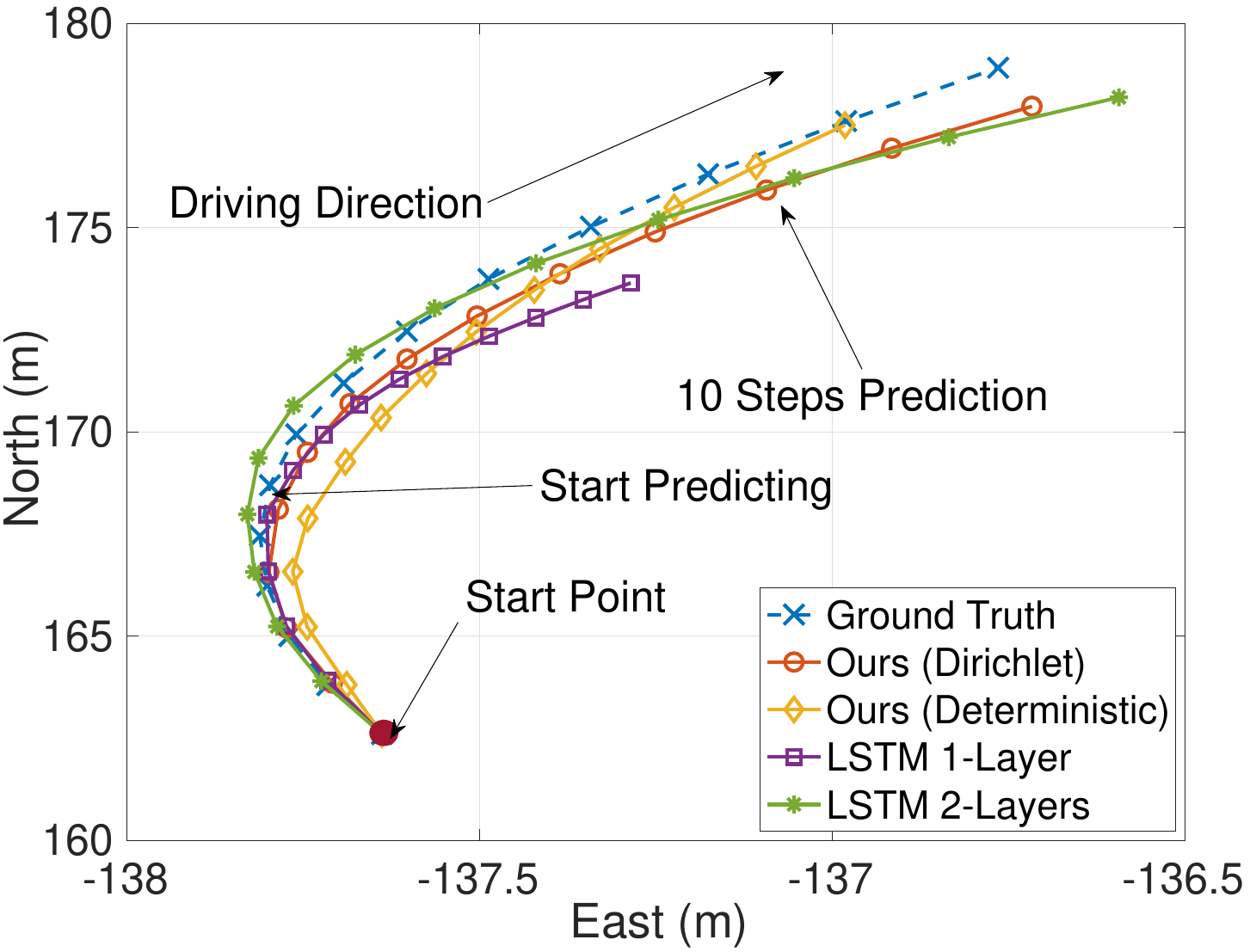}
        	\caption{\label{fig: turning inertial good} Turning pred. with inertial data in the good case}
        \end{subfigure}
        \begin{subfigure}[t]{0.32\textwidth}
    	\includegraphics[width=\textwidth]{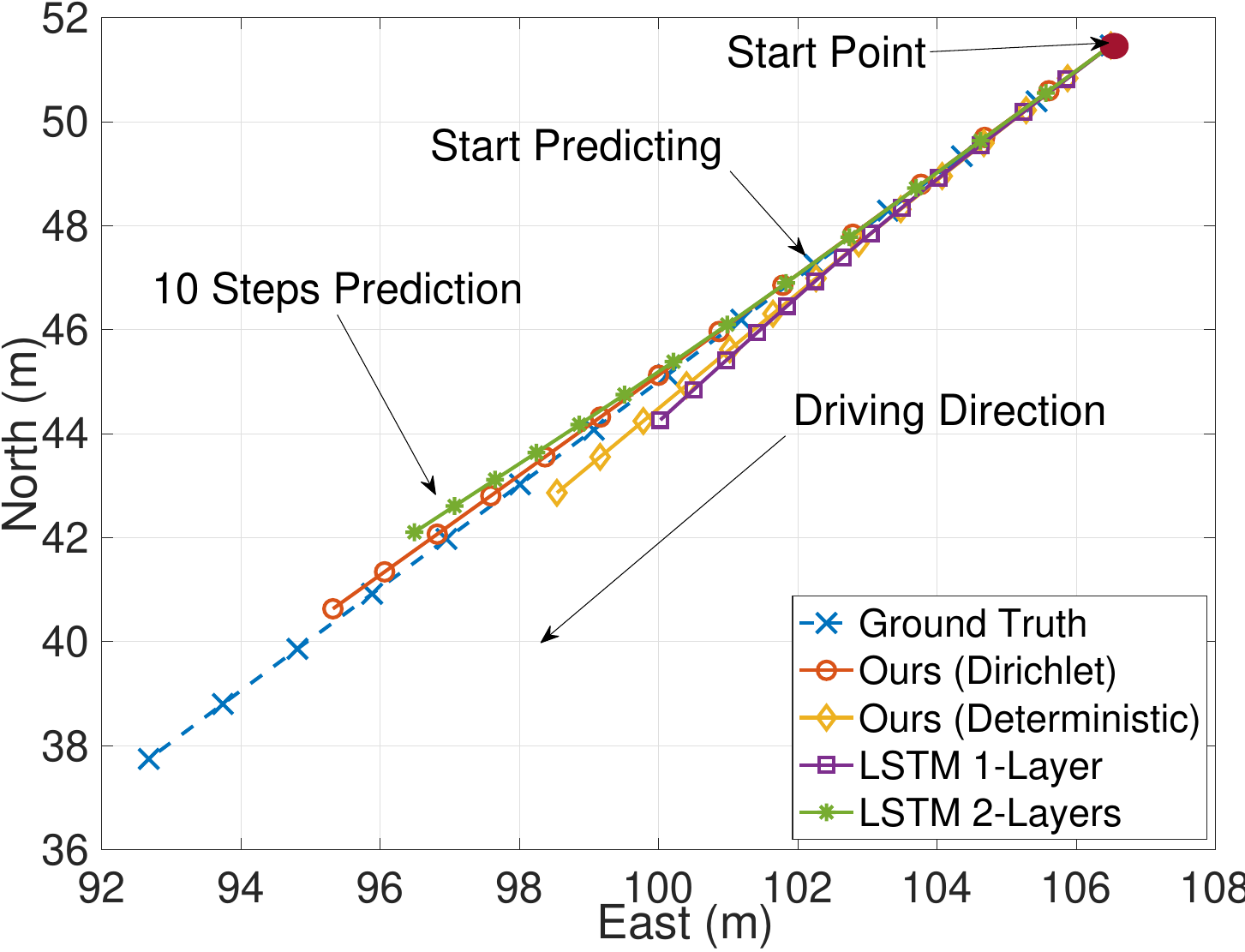}
        	\caption{\label{fig: line worst} Line prediction in the bad case}
        \end{subfigure}
         \begin{subfigure}[t]{0.32\textwidth}
    	\includegraphics[width=\textwidth]{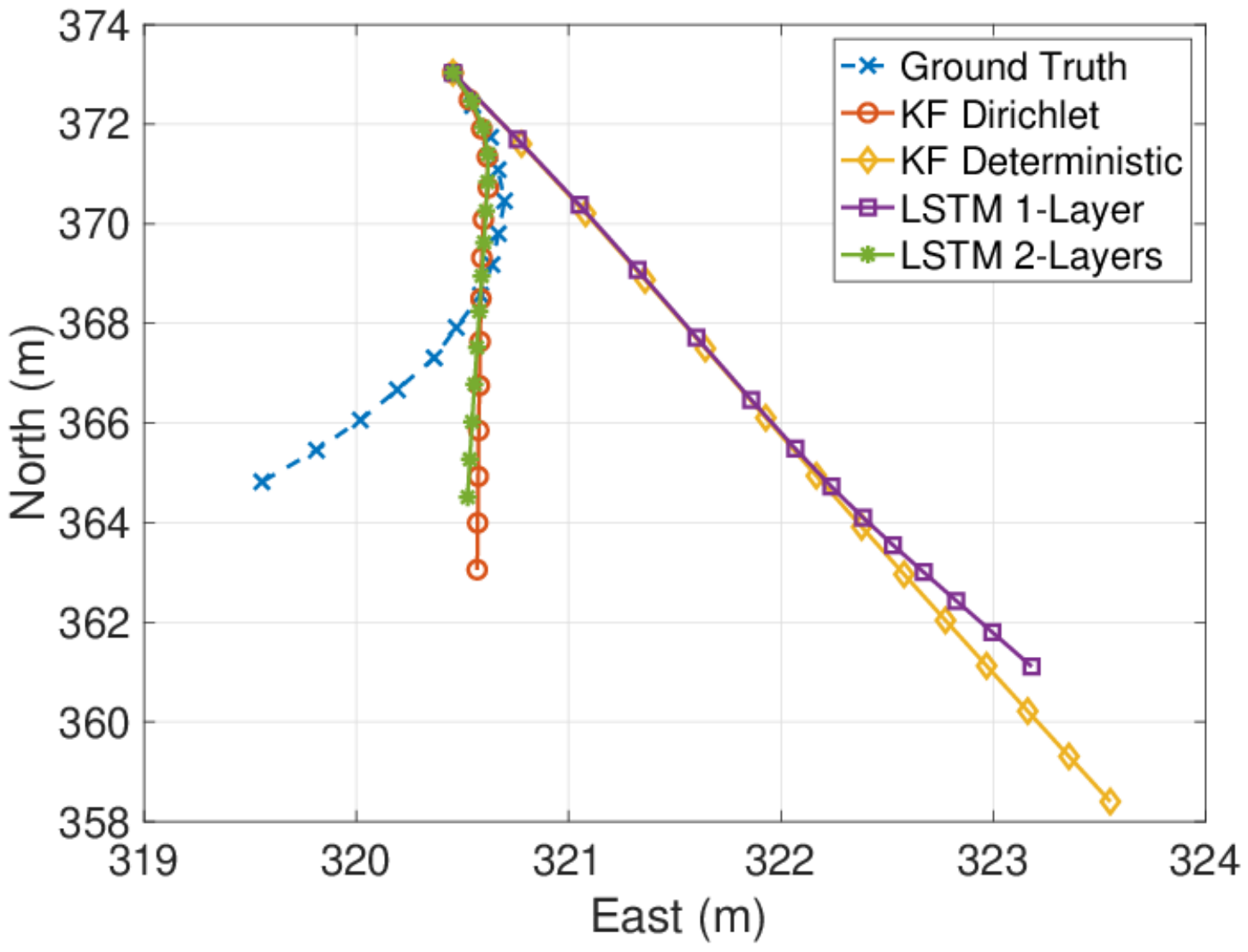}
        	\caption{\label{fig: turning bad} Turning pred. w/o inertial data in the bad case}
        \end{subfigure}
        \begin{subfigure}[t]{0.32\textwidth}
        \includegraphics[width=\textwidth]{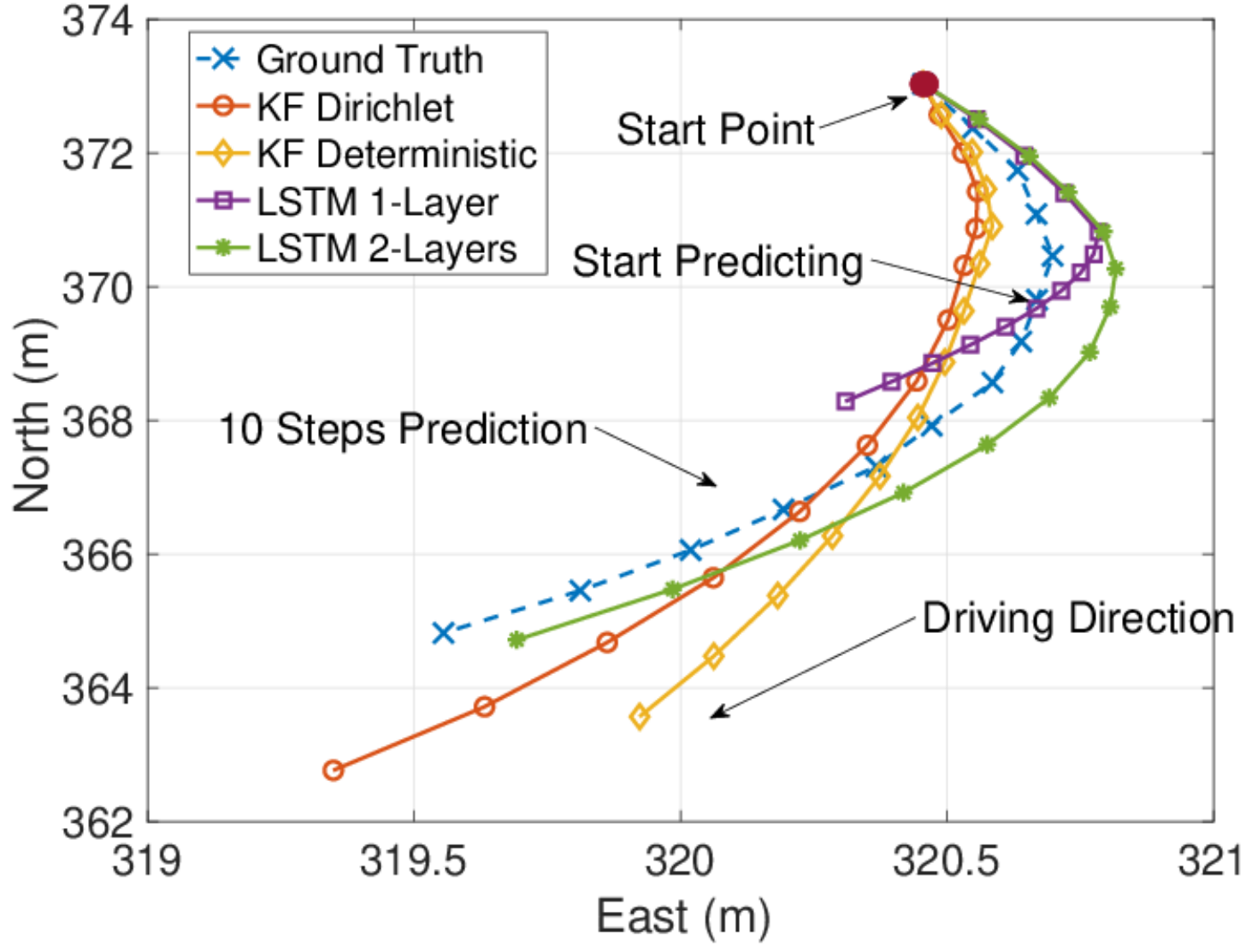}
        	\caption{\label{fig: turning inertial bad} Turning pred. with inertial data in the bad case}
        \end{subfigure}
        \caption{\label{fig: turn prediction} For future poses prediction without observations, our Dirichlet based model clearly outperforms others when predicting the straight driving. In both (a) the good case and (d) the bad case, the predicted locations in future steps from our proposed Dirichlet based DynaNet model are closer to the groundtruth, compared with other baselines. In turning, the future poses are estimated in a tangent direction with or without the aid of inertial data (b) (c) in good case and (e) (f) bad case. }
    \end{figure*}
    
        \begin{figure}
    \includegraphics[width=0.45\textwidth]{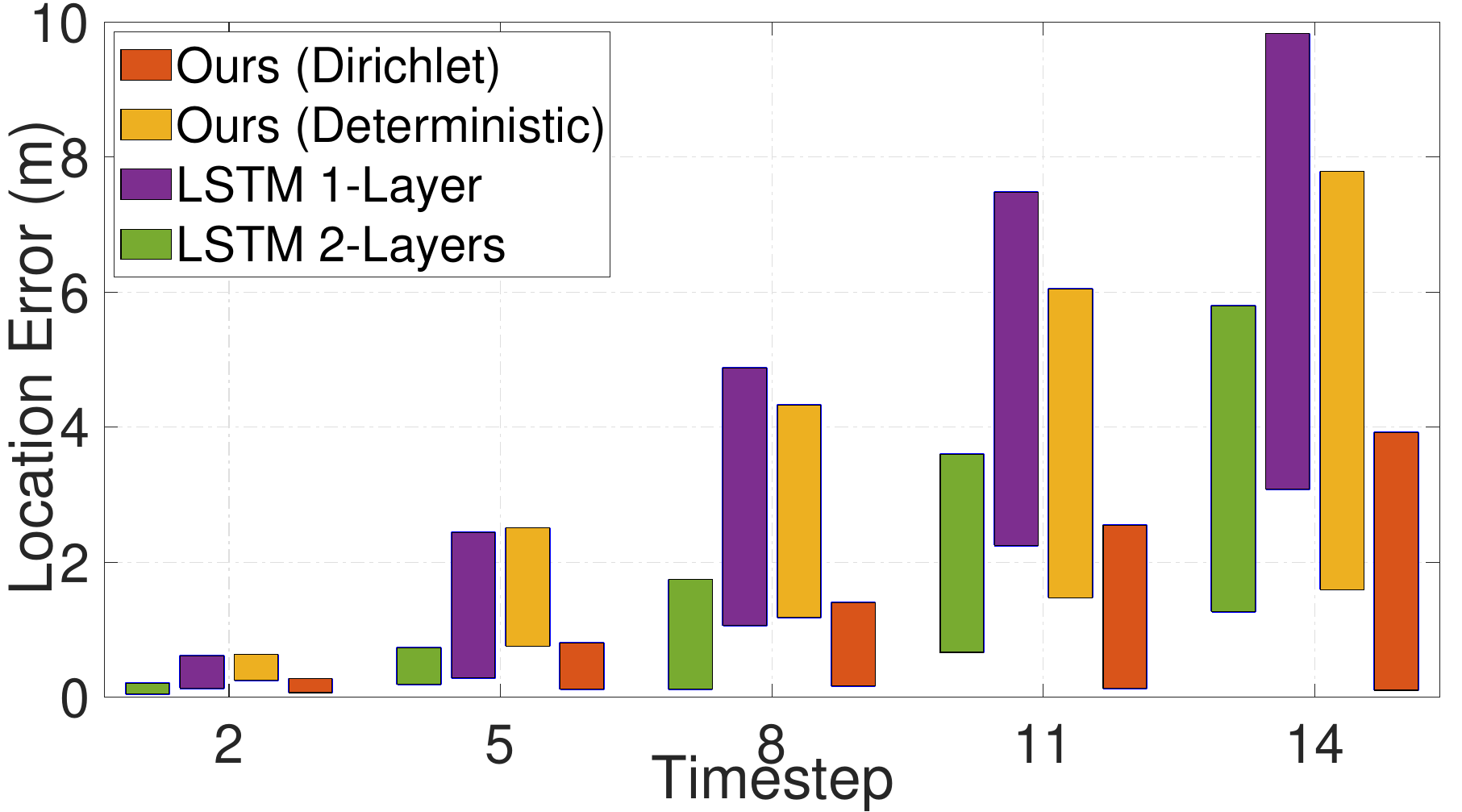}
    \captionof{figure}{ The error bar of line predictions
         }
    \label{fig: error line}
\end{figure}

\subsubsection{VIO Pose Estimation}
In this experiment, we adopt the official KITTI evaluation metric to evaluate our models and baseline, which is the same as in VO experiments.
Table \ref{tab: vio} reports the RMSE of the translation and orientation of the proposed DynaNet models, {a classical model based VIO algorithm, i.e. VINS-Mono \cite{Qin2018}} and a learning based approach, i.e. VINet \cite{Clark2017a}. 
{Due to the problem of loosely time-synchronization between visual and inertial sensors in KITTI dataset, the performance of VINS-mono is not as good as learning-based methods.
This supports the claim that learning-based approaches perform more robustly than hand-designed systems.}
From Table II, our proposed models outperform VINet with 2-layers LSTMs, when given both visual and inertial observations. VINet shares the same framework and hyperparameters as our models, except that it uses LSTM rather than a differentiable Kalman filtering. Our deterministic DynaNet (Ours (Deter.)) further reduces the RMSE of VINet from 6.44\% to 5.47\% in translation, and from 1.70$^{\circ}$ to 1.63$^{\circ}$.
This demonstrates that our proposed models excel at fusing multiple sensor modalities for more accurate state estimates than the LSTM based VIO model.

\subsection{Motion Prediction}
\label{sub: motion prediction}
In this experiment, we show the evaluation of DynaNet models on pose prediction that offers pose without sensor data (i.e. future states prediction).

\subsubsection{VO Pose Prediction}
We first fed VO neural models a sequence of 5 images for initialisation, and then let them predict the next 5 and 10 states without any further observations (i.e. trajectory prediction). All models including baselines were trained on the training set of the KITTI dataset, and tested on the sub-sequences of 10 or 15 frames, generated from the testing set. In order to compare with LSTM baselines fairly, the structures and hyperparameters of baseline models are kept the same as our models except the DynaNet module.

Table \ref{tab: motion prediction} illustrates the quantitative results of our approaches, comparing with LSTM based and {``LSTM+Attention" based} models. We report the RMSE of relative positions of the next 5 and 10 steps predicted by neural networks. 
As shown in Table \ref{tab: motion prediction}, it is clear that our proposed models perform better than both LSTM based and {``LSTM+Attention" based} models in visual egomotion prediction. 
Especially, our Dirichlet model outperforms others by a large margin. This is because the resampled transition matrix from the Dirichlet distribution ensures the learned dynamical model to be stable, and hence gives rise to long-term prediction in higher accuracy. 

 \begin{figure*}
    \includegraphics[width=0.95\textwidth]{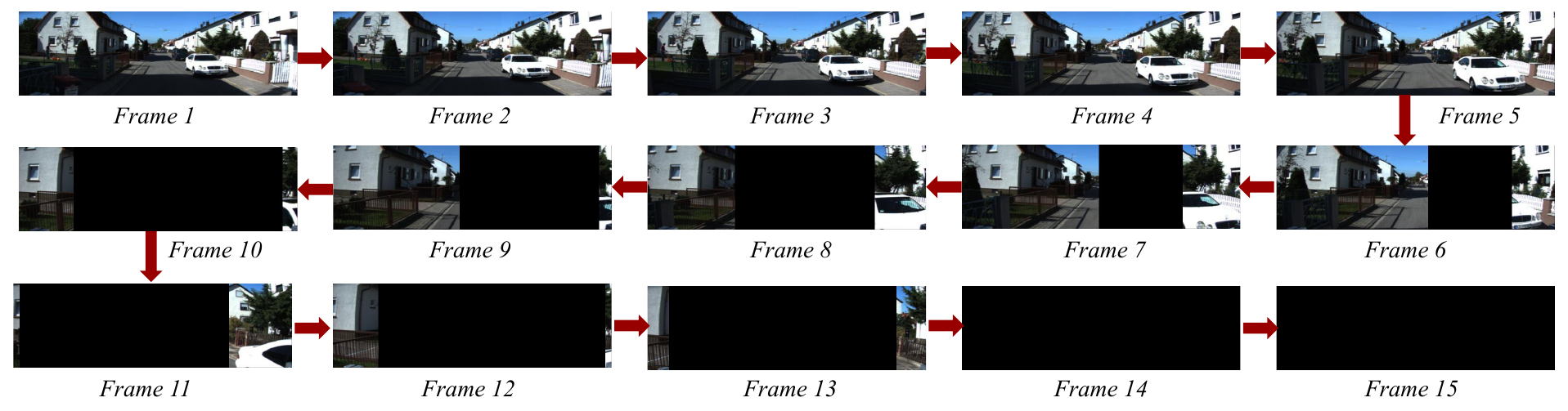}
    \captionof{figure}{ Sample images from the generated sub-sequences degraded with increasing size of blanked block in the interpretability experiment.}
    \label{fig: block}
\end{figure*}

 	\begin{figure*}
    	\centering
        \begin{subfigure}[t]{0.42\textwidth}
        	\includegraphics[width=\textwidth]{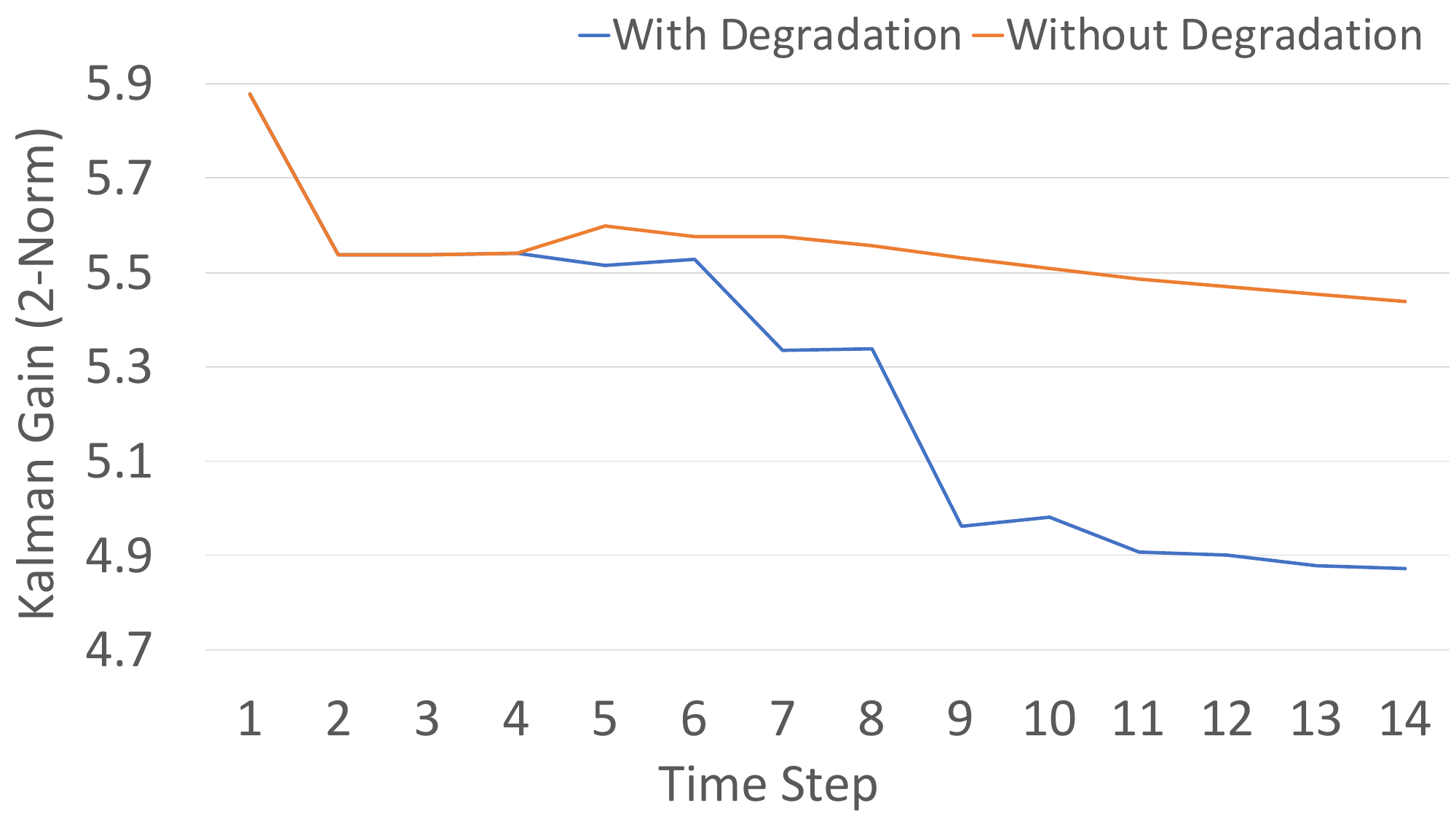}
        	\caption{\label{fig: kalman gain} Kalman Gain}
        \end{subfigure}
        \begin{subfigure}[t]{0.42\textwidth}
        	\includegraphics[width=\textwidth]{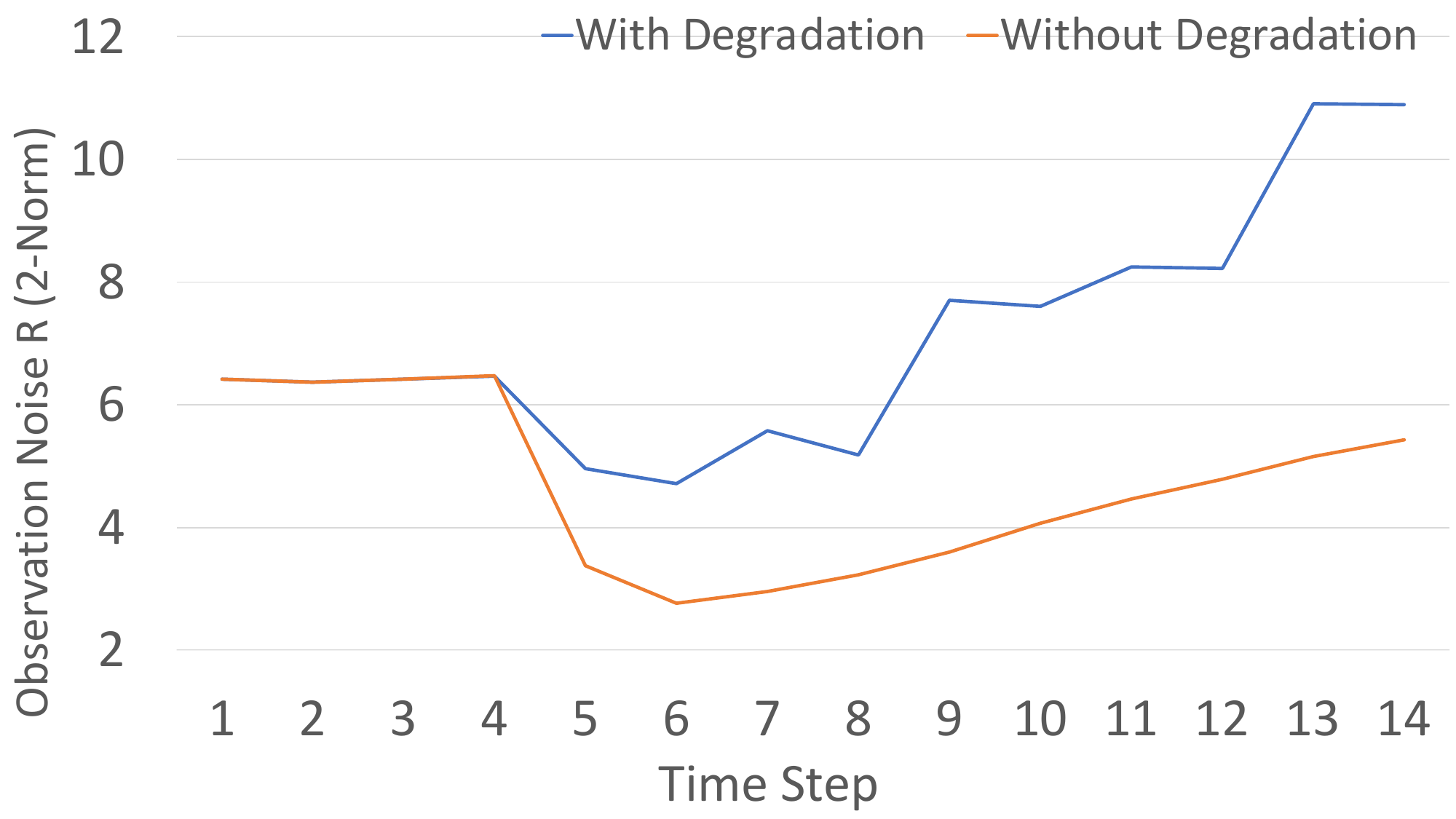}
        	\caption{\label{fig: obs} Observation Noise}
        \end{subfigure}
        \begin{subfigure}[t]{0.42\textwidth}
        	\includegraphics[width=\textwidth]{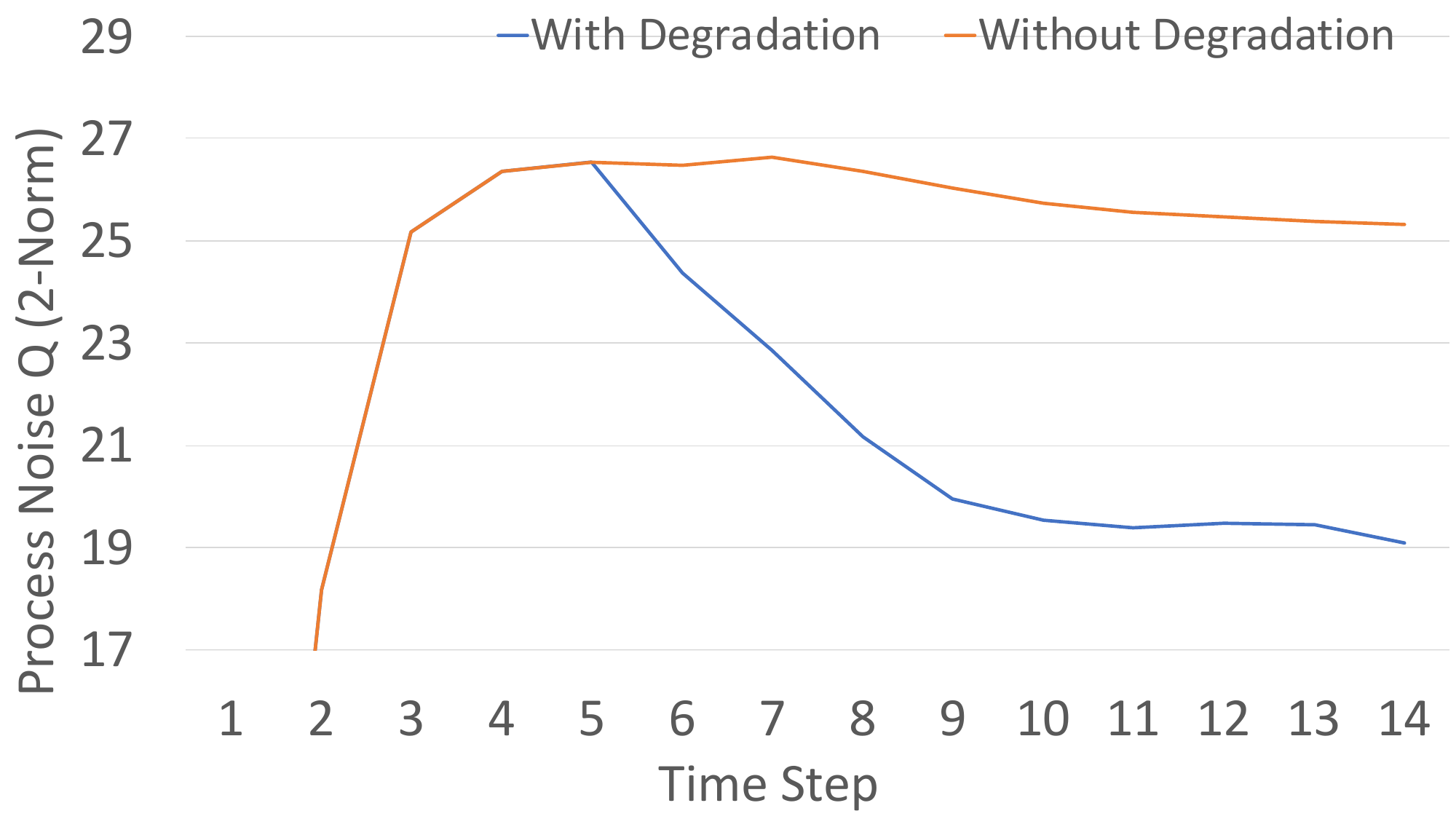}
        	\caption{\label{fig: process} Process Noise}
        \end{subfigure}
        \begin{subfigure}[t]{0.42\textwidth}
        	\includegraphics[width=\textwidth]{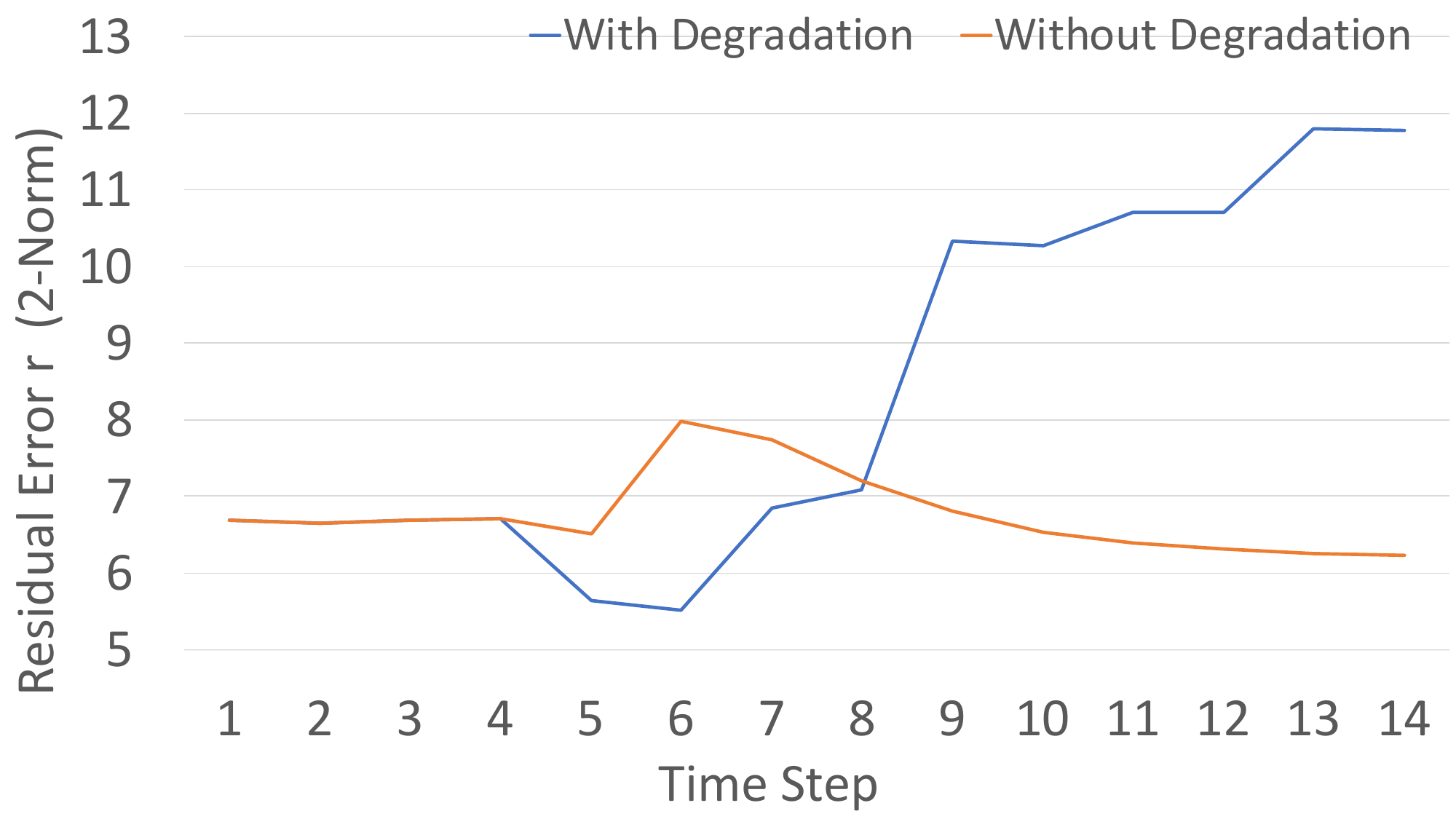}
        	\caption{\label{fig: res} Residual Noise}
        \end{subfigure}
        \caption{\label{fig: trajectories} Model Interpretation. The Kalman gain (a) reflects the measurement qualities which decreases with the rising degree of data corruption. The increasing observation noise (b) and residual noise (d) reflect the rise of observation uncertainty. The decreasing process noise (d) indicates that the model places more trust in process prediction when observations are uncertain.}
    \end{figure*}

We then demonstrate the qualitative results of our methods.
As straight driving routes dominate driving behaviours for autonomous vehicles, we thus begin our performance report with them first. 
Figure \ref{fig: line best} and \ref{fig: line worst} show the best case and worst case of line prediction. Clearly, in both cases, the predicted trajectories from our proposed Dirichlet model are much closer to the groundtruth trajectories. Without observations, in the best case it can even provide the same predictions (the small circles in Figure \ref{fig: line best}) as groundtruth values (the small crosses).
Figure \ref{fig: error line} plots the error bar of predicted locations for straight driving, with the best and worst results among all tested segments to show the variance of the model predictions. 
In both cases, our proposed Dirichlet based model consistently outperforms other baselines by providing more accurate and robust location predictions. 

Figure \ref{fig: turning good} and \ref{fig: turning bad} further show the prediction performance in turning. As we can see, without timely observation, it is hard for deep neural networks to estimate accurate orientation changes, but they predict the future poses in a tangent direction in both good and bad cases. In the bad case, the motion predictions are relatively more far away from the ground-truth.
We will soon show how to integrate inertial information to aid the turning prediction in VIO pose prediction.

\subsubsection{VIO Pose Prediction}
We also evaluate our models on pose prediction using visual and inertial sensor data. 
This experiment is conducted in scenarios of prediction with visual-only observations, inertial-only observations and no observations, in which all models are given a sequence of 5 images for initialisation and need to predict the next 5 poses. 

As can be seen in Table \ref{tab: vio prediction}, our models including deterministic and Dirichlet approaches, greatly outperform the comparable LSTM {and ``LSTM+Attention"} approaches. 
{When no input is available, although attention mechanism improves the prediction performance over LSTM, our DynaNet still outperforms this ``LSTM+attention" baseline. Specifically, the Dirichlet model shows better performance than all other approaches.} This is consistent with the results in visual pose prediction, and validates that the model stability will allow more accurate long term prediction.

Figure \ref{fig: turning inertial good} and \ref{fig: turning inertial bad} plot the predicted trajectories with only inertial data, when no visual observation is given. Clearly, in the good case our models are capable of predicting future pose evolution accurately and robustly. We note that robust fusion is important for safe operation with missing sensor inputs e.g. for self-driving cars. In the bad case, though the results are not desirable, the motion predictions still indicate the tendency of turning.

\subsection{Towards Model Interpretability} 
\label{sub:kalman_gain}
We are now in a position to discuss model interpretability. Recall that in the update process (see Section~\ref{sub:prediction_and_inference_via_kalman_filter}), the Kalman gain is an adaptive weight that balances the observations and the model predictions. If there is a high confidence in measurements, the Kalman gain will increase to selectively upweight measurement innovation and vice versa. This property gives us a unique opportunity to analyse model behaviours from the value changes of the Kalman Gain.

To this end, we deliberately fed our Dirichlet model with degraded images and use Kalman gain to capture the belief in measurements. This experiment generated 113 sequences with 15 frames of images from Sequence 09 in KITTI dataset. For data degradation, a block was blanked on a sequence of 15 frames of images. The size of blank blocks gradually increases as time evolves, until all pixels on an image are blank. Specifically, as shown in Figure \ref{fig: block}, in each sequence, the images are corrupted with an increasing size of blanked blocks on the timesteps 1-5 (no blanked block), 6-7 (a blanked block with 192 pixels $\times$ 192 pixels), 8-9 (a blanked block with 192 pixels  $\times$ 320 pixels), 10-11 (a blanked block with 192 pixels $\times$ 480 pixels), 12-13 (a blanked block with 192 pixels $\times$ 520 pixels), 14-15 (a blanked block with 192 pixels $\times$ 640 pixels). The position of the blanked block is randomly selected on each image. 
We then test our model with this modified dataset in the same fashion as the VO experiment described in Section~\ref{sub:visual_egomotion_estimation}.

The Frobenius norm (2-norm) of the Kalman gain matrix is calculated, and averaged across all sequences as an aggregated indicator of changes in Kalman gain matrix. 
Figure \ref{fig: kalman gain} shows that compared to the case with no degradation, this indicator gradually decreases with growing data corruption. It implies that our model can adaptively place more trust on model predictions when observing low-quality data and signal to higher control layers that estimation is becoming more uncertain. Similarly, we further visualize other explicit parameters inside our DynaNet model, i.e. the observation noise matrix $\mathbf{R}$, process noise matrix $\mathbf{Q}$ and residual noise matrix $\mathbf{r}$. As shown in Figure \ref{fig: obs}, \ref{fig: process} and \ref{fig: res}, these parameters are also able to capture and reflect the model uncertainty: with respect to the increasing data corruption, the observation noise rises up as well, indicating that the model are more uncertain about the observations. On the contrary, the process noise goes down, showing that the model has to place more belief in prediction process when the observations are uncertain. Last but not least, the increasing residual noise also aligns with the uncertainty in observations. \textit{It is critical to note that data corrupted in this way have never been seen in the training phase.}

\section{Conclusion}
DynaNet, a neural Kalman dynamical model was introduced in this paper to learn temporary linear-like structure on latent states. Through deeply coupled DNNs and SSMs, DynaNet can scale to high-dimensional data as well as model very complex motion dynamics in real world. By using Kalman filter on feature space, DynaNet is able to reason about latent system states, allowing reliable inference and predictions even with missing observations. Furthermore, the transition matrix in our model is sampled from the Dirichlet distribution learned by a RNN, which ensures system stability in the long run. 
DynaNet is evaluated on a variety of challenging motion-estimation tasks, including single-modality estimation under data corruption, multiple sensor fusion under data absence, and future motion prediction. Experimental results demonstrate the superiority of our approach in accuracy, robustness and interpretability.

{In the future work, it would be interesting to further explore the application of proposed DynaNet model in motion prediction, e.g. evaluating and visualizing future steps predictions in various environments, and measuring the robustness and reliability of a learned stable dynamical model compared with an unstable model.
}

\appendices
\section{The implementation details of DynaNet models}
This appendix illustrates the implementation details of the experiments in visual odometry and visual-inertial odometry.
All the models are trained with 100 epochs. 

\textbf{Visual Odometry:} Table \ref{tab: vo framework} reports the framework for the visual odometry,  consisting of Visual Encoder to extract features $\mathbf{a}$ and observation noise matrix $\mathbf{Q}$, Neural Transition Models (Deterministic or Resampled) to generate transition matrix $\mathbf{A}$ and process noise matrix $\mathbf{R}$, the Kalman Filter pipeline to predict and update system states $\mathbf{z}$, and Pose Predictor to output 6-states poses ([Translation, Euler angles]) from systems states.
Specifically, our visual encoder used the encoder structure of the FlowNetS architecture. 

\begin{table}[H]
\centering
\resizebox{\columnwidth}{!}{
\begin{tabular}{l}
Visual Encoder \\
\hline 
$\lbrack$ input $\rbrack$ Two stacked images: $B\times640\times192\times6$   \\     
$\lbrack$ layer 1 $\rbrack$ Conv. $7^2$, Stride $2^2$, Padding 3, LeakyReLU activ.\\
$\lbrack$ layer 2 $\rbrack$ Conv. $5^2$, Stride $2^2$, Padding 2, LeakyReLU activ.\\
$\lbrack$ layer 3 $\rbrack$ Conv. $5^2$, Stride $2^2$, Padding 2, LeakyReLU activ.\\
$\lbrack$ layer 3\_1 $\rbrack$ Conv. $3^2$, Stride $1^2$, Padding 1 \\
$\lbrack$ layer 4 $\rbrack$ Conv. $3^2$, Stride $2^2$, Padding 2, LeakyReLU activ.\\
$\lbrack$ layer 4\_1 $\rbrack$ Conv. $3^2$, Stride $1^2$, Padding 1 \\
$\lbrack$ layer 5 $\rbrack$ Conv. $3^2$, Stride $2^2$, Padding 2, LeakyReLU activ.\\
$\lbrack$ layer 5\_1 $\rbrack$ Conv. $3^2$, Stride $1^2$, Padding 1 \\
$\lbrack$ layer 6 $\rbrack$ Conv. $3^2$, Stride $2^2$, Padding 1 \\
$\lbrack$ layer 7\_1 $\rbrack$ FC 128\\
$\lbrack$ layer 7\_2 $\rbrack$ FC 128, ReLU\\
$\lbrack$ output1 $\rbrack$ Observation $\mathbf{a}$: $B\times128$ (layer 7\_1)\\
$\lbrack$ output2 $\rbrack$ Observation Noise Covariance $\mathbf{R}$: $B\times128$ (layer 7\_2) \\
\\

Neural Transition Generation - Deterministic\\
\hline
$\lbrack$ input $\rbrack$ Latent States $z$: $B\times1\times128$   \\
$\lbrack$ layer 1 $\rbrack$ LSTM, 1-layer, hidden size 128\\
$\lbrack$ layer 2\_1 $\rbrack$ FC 128\\
$\lbrack$ layer 2\_2 $\rbrack$ FC 128, ReLU\\
$\lbrack$ output 1 $\rbrack$ Transition Matrix $\mathbf{A}$: $B\times128$ (layer 2\_1) \\
$\lbrack$ output 2 $\rbrack$ Process Noise Covariance $\mathbf{Q}$: $B\times128$ (layer 2\_2) \\
\\

Neural Transition Generation - Resampled\\
\hline
$\lbrack$ input $\rbrack$ Latent States z: $B\times1\times128$   \\
$\lbrack$ layer 1 $\rbrack$ LSTM, 1-layer, hidden size 128\\
$\lbrack$ layer 2\_1 $\rbrack$ FC 128, ReLU\\
$\lbrack$ layer 2\_2 $\rbrack$ FC 128, ReLU\\
$\lbrack$ layer 3 $\rbrack$ Dirichlet Distribution\\
$\lbrack$ output 1 $\rbrack$ Transition Matrix $\mathbf{A}$: $B\times128$ (layer3)\\
$\lbrack$ output 2 $\rbrack$ Process Noise Covariance $\mathbf{Q}$: $B\times128$ (layer 2\_2)\\
\\

Kalman Filter Pipeline\\
\hline
$\lbrack$ input1 $\rbrack$ Previous Latent States $\mathbf{z}_{t-1}$: $B\times128$   \\
$\lbrack$ input2 $\rbrack$ Previous State Covariance $\mathbf{P}_{t-1}$: $B\times128$   \\
$\lbrack$ input3 $\rbrack$ Transition $\mathbf{A}$ and Process Noise $\mathbf{Q}$: $B\times128$, $B\times128$   \\
$\lbrack$ input4 $\rbrack$ Observation $\mathbf{a}$ and Observation Noise $\mathbf{R}$: $B\times128$, $B\times128$   \\
$\lbrack$ layer 1 $\rbrack$ Kalman Filter\\
$\lbrack$ output 1 $\rbrack$ Current Latent States $\mathbf{z}_{t}$: $B\times128$ \\
$\lbrack$ output 2 $\rbrack$ Current State Covariance $\mathbf{P}_{t}$: $B\times128$ \\
\\

Pose Predictor \\
\hline
$\lbrack$ input $\rbrack$ Current Latent States $\mathbf{z}_{t}$: $B\times128$   \\     
$\lbrack$ layer 1\_1 $\rbrack$  FC 3\\
$\lbrack$ layer 1\_2 $\rbrack$ FC 3\\
$\lbrack$ layer 4 $\rbrack$ (layer 1\_1) $\oplus$ (layer 1\_2)  \\
$\lbrack$ output $\rbrack$ Poses: $B\times6$ \\
\end{tabular}
}

\caption{Implementation details for the Visual Odometry Experiment.}
\label{tab: vo framework}
\end{table}

\textbf{Visual-Inertial Odometry} As can be seen in Table \ref{tab: vio framework}, the framework for visual-inertial odometry used the same Neural Transition Model, Kalman Filter, and Pose Predictor as in visual odometry, except the encoders and sensor fusion part. Table 2 shows that the  64-dimensional visual and inertial features are extracted from data by visual and inertial encoders respectively, and concatenated as a 128-dimensional observation in Sensor Fusion. The inertial encoder used 1-layer Bi-directional LSTM to process inertial data.

\begin{table}[H]
\centering
\resizebox{\columnwidth}{!}{
\begin{tabular}{l}
Visual Encoder \\
\hline 
$\lbrack$ input $\rbrack$ Two stacked images: $B\times640\times192\times6$   \\     
$\lbrack$ layer 1 $\rbrack$ Conv. $7^2$, Stride $2^2$, Padding 3, LeakyReLU activ.\\
$\lbrack$ layer 2 $\rbrack$ Conv. $5^2$, Stride $2^2$, Padding 2, LeakyReLU activ.\\
$\lbrack$ layer 3 $\rbrack$ Conv. $5^2$, Stride $2^2$, Padding 2, LeakyReLU activ.\\
$\lbrack$ layer 3\_1 $\rbrack$ Conv. $3^2$, Stride $1^2$, Padding 1 \\
$\lbrack$ layer 4 $\rbrack$ Conv. $3^2$, Stride $2^2$, Padding 2, LeakyReLU activ.\\
$\lbrack$ layer 4\_1 $\rbrack$ Conv. $3^2$, Stride $1^2$, Padding 1 \\
$\lbrack$ layer 5 $\rbrack$ Conv. $3^2$, Stride $2^2$, Padding 2, LeakyReLU activ.\\
$\lbrack$ layer 5\_1 $\rbrack$ Conv. $3^2$, Stride $1^2$, Padding 1 \\
$\lbrack$ layer 6 $\rbrack$ Conv. $3^2$, Stride $2^2$, Padding 1 \\
$\lbrack$ layer 7\_1 $\rbrack$ FC 64\\
$\lbrack$ layer 7\_2 $\rbrack$ FC 64, ReLU\\
$\lbrack$ output1 $\rbrack$ Visual Feature $\mathbf{a}_v$: $B\times64$ (layer 7\_1)\\
$\lbrack$ output2 $\rbrack$ Visual Noise Covariance $\mathbf{R}$: $B\times64$ (layer 7\_2) \\
\\

Inertial Encoder \\
\hline 
$\lbrack$ input $\rbrack$ IMU sequence: $B\times10\times6$   \\
$\lbrack$ layer 1 $\rbrack$ FC 32\\
$\lbrack$ layer 2 $\rbrack$ B-LSTM, 2-layers, hidden size 32\\
$\lbrack$ layer 2\_1 $\rbrack$ FC 64\\
$\lbrack$ layer 2\_2 $\rbrack$ FC 64, ReLU\\
$\lbrack$ output1 $\rbrack$ Inertial Feature $\mathbf{a}_i$: $B\times64$ (layer 2\_1) \\
$\lbrack$ output2 $\rbrack$ Inertial Noise Covariance $\mathbf{R}$: $B\times64$ (layer 2\_2)\\
\\

Sensor Fusion \\
\hline 
$\lbrack$ input1 $\rbrack$ Visual Feature $\mathbf{a}_v$: $B\times64$   \\
$\lbrack$ input2 $\rbrack$ Inertial Feature $\mathbf{a}_i$: $B\times64$   \\
$\lbrack$ layer 1 $\rbrack$ $(\text{input1} \oplus \text{input2})$\\
$\lbrack$ output $\rbrack$ Observation $\mathbf{a}$: $B\times128$ \\

\end{tabular}
}

\caption{Implementation details for Visual Inertial Odometry. $\oplus$ denotes a concatenation operation.}
\label{tab: vio framework}
\end{table}

\bibliographystyle{IEEEtran}
\bibliography{refs.bib}

\begin{thebibliography}{10}
\providecommand{\url}[1]{#1}
\csname url@samestyle\endcsname
\providecommand{\newblock}{\relax}
\providecommand{\bibinfo}[2]{#2}
\providecommand{\BIBentrySTDinterwordspacing}{\spaceskip=0pt\relax}
\providecommand{\BIBentryALTinterwordstretchfactor}{4}
\providecommand{\BIBentryALTinterwordspacing}{\spaceskip=\fontdimen2\font plus
\BIBentryALTinterwordstretchfactor\fontdimen3\font minus
  \fontdimen4\font\relax}
\providecommand{\BIBforeignlanguage}[2]{{%
\expandafter\ifx\csname l@#1\endcsname\relax
\typeout{** WARNING: IEEEtran.bst: No hyphenation pattern has been}%
\typeout{** loaded for the language `#1'. Using the pattern for}%
\typeout{** the default language instead.}%
\else
\language=\csname l@#1\endcsname
\fi
#2}}
\providecommand{\BIBdecl}{\relax}
\BIBdecl

\bibitem{Sunderhauf2018}
N.~S{\"{u}}nderhauf, O.~Brock, W.~Scheirer, R.~Hadsell, D.~Fox, J.~Leitner,
  B.~Upcroft, P.~Abbeel, W.~Burgard, M.~Milford, and P.~Corke, ``{The Limits
  and Potentials of Deep Learning for Robotics},'' \emph{International Journal
  of Robotics Research}, vol.~37, no. 4-5, pp. 405--420, 2018.

\bibitem{Nister2004}
D.~Nister, O.~Naroditsky, and J.~Bergen, ``{Visual Odometry},'' in \emph{IEEE
  Conference on Computer Vision and Pattern Recognition (CVPR)}, vol.~1, 2004,
  pp. I--652--I--659 Vol.1.

\bibitem{Engel2013}
J.~Engel, J.~Sturm, and D.~Cremers, ``{Semi-Dense Visual Odometry for a
  Monocular Camera},'' in \emph{IEEE International Conference on Computer
  Vision (ICCV)}, 2013, pp. 1449--1456.

\bibitem{Forster2014}
C.~Forster, M.~Pizzoli, and D.~Scaramuzza, ``{SVO: Fast Semi-Direct Monocular
  Visual Odometry},'' in \emph{IEEE International Conference on Robotics and
  Automation (ICRA)}, 2014, pp. 15--22.

\bibitem{Kalman1960}
R.~E. Kalman, ``{A New Approach to Linear Filtering and Prediction Problems},''
  \emph{Journal of Basic Engineering}, vol.~82, no.~1, p.~35, 1960.

\bibitem{Mohinder2010}
S.~G. Mohinder and P.~A. Angus, ``{Applications of Kalman Filtering in
  Aerospace 1960 to the Present},'' \emph{IEEE Control Systems Magazine}, pp.
  69--78, 2010.

\bibitem{Liu1998}
J.~S. Liu and R.~Chen, ``Sequential monte carlo methods for dynamic systems,''
  \emph{Journal of the American Statistical Association}, vol.~93, no. 443, pp.
  1032--1044, 1998.

\bibitem{Kummerle2011}
R.~K{\"{u}}mmerle, G.~Grisetti, H.~Strasdat, K.~Konolige, and W.~Burgard,
  ``{g2o: A General Framework for Graph Optimization},'' in \emph{IEEE
  International Conference on Robotics and Automation (ICRA)}, 2011.

\bibitem{Clark2017}
R.~Clark, S.~Wang, A.~Markham, N.~Trigoni, and H.~Wen, ``{VidLoc: A Deep
  Spatio-Temporal Model for 6-DoF Video-Clip Relocalization},'' in \emph{IEEE
  Conference on Computer Vision and Pattern Recognition (CVPR)}, 2017.

\bibitem{Wang2017}
S.~Wang, R.~Clark, H.~Wen, and N.~Trigoni, ``{DeepVO : Towards End-to-End
  Visual Odometry with Deep Recurrent Convolutional Neural Networks},'' in
  \emph{IEEE International Conference on Robotics and Automation (ICRA)}, 2017.

\bibitem{Zhou2017}
T.~Zhou, M.~Brown, N.~Snavely, and D.~G. Lowe, ``{Unsupervised Learning of
  Depth and Ego-Motion from Video},'' in \emph{IEEE Conference on Computer
  Vision and Pattern Recognition (CVPR)}, 2017.

\bibitem{Clark2017a}
R.~Clark, S.~Wang, H.~Wen, A.~Markham, and N.~Trigoni, ``{VINet:
  Visual-Inertial Odometry as a Sequence-to-Sequence Learning Problem},'' in
  \emph{Association for the Advancement of Artificial Intelligence (AAAI)},
  2017, pp. 3995--4001.

\bibitem{Mirowski2018}
P.~Mirowski, M.~K. Grimes, M.~Malinowski, K.~M. Hermann, K.~Anderson,
  D.~Teplyashin, K.~Simonyan, K.~Kavukcuoglu, A.~Zisserman, and R.~Hadsell,
  ``{Learning to Navigate in Cities Without a Map},'' in \emph{Advances in
  Neural Information Processing Systems (NIPS)}, 2018.

\bibitem{Bloesch2018}
M.~Bloesch, J.~Czarnowski, R.~Clark, S.~Leutenegger, and A.~J. Davison,
  ``{CodeSLAM — Learning a Compact, Optimisable Representation for Dense
  Visual SLAM},'' in \emph{IEEE Conference on Computer Vision and Pattern
  Recognition (CVPR)}, 2018.

\bibitem{Brahmbhatt2018}
S.~Brahmbhatt, J.~Gu, K.~Kim, J.~Hays, and J.~Kautz, ``{Geometry-Aware Learning
  of Maps for Camera Localization},'' in \emph{IEEE Conference on Computer
  Vision and Pattern Recognition (CVPR)}, 2018, pp. 2616--2625.

\bibitem{Zhan2018}
\BIBentryALTinterwordspacing
H.~Zhan, R.~Garg, C.~S. Weerasekera, K.~Li, H.~Agarwal, and I.~Reid,
  ``{Unsupervised Learning of Monocular Depth Estimation and Visual Odometry
  with Deep Feature Reconstruction},'' in \emph{IEEE Conference on Computer
  Vision and Pattern Recognition (CVPR)}, 2018, pp. 340--349. [Online].
  Available: \url{http://arxiv.org/abs/1803.03893}
\BIBentrySTDinterwordspacing

\bibitem{Yin2018}
Z.~Yin and J.~Shi, ``{GeoNet: Unsupervised Learning of Dense Depth, Optical
  Flow and Camera Pose},'' in \emph{IEEE Conference on Computer Vision and
  Pattern Recognition (CVPR)}, 2018.

\bibitem{Dudley1979}
D.~G. {Dudley}, ``Dynamic system identification experiment design and data
  analysis,'' \emph{Proceedings of the IEEE}, vol.~67, no.~7, pp. 1087--1087,
  July 1979.

\bibitem{yu2017incremental}
J.~Yu, S.~Wang, and L.~Li, ``Incremental design of simplex basis function model
  for dynamic system identification,'' \emph{IEEE transactions on neural
  networks and learning systems}, vol.~29, no.~10, pp. 4758--4768, 2017.

\bibitem{Kocijan2005}
J.~Kocijan, A.~Girard, B.~Banko, and R.~Murray-Smith, ``{Dynamic Systems
  Identification with Gaussian Processes},'' \emph{Mathematical and Computer
  Modelling of Dynamical Systems}, vol.~11, no.~4, pp. 411--424, 2005.

\bibitem{Ghahramani1999}
Z.~Ghahramani and S.~T. Roweis, ``{Learning Nonlinear Dynamical Systems Using
  an EM Algorithm},'' in \emph{Advances in Neural Information Processing
  Systems (NIPS)}, vol.~11, no.~1, 1999, pp. 431--437.

\bibitem{wang2011convergence}
X.~Wang and Y.~Huang, ``Convergence study in extended kalman filter-based
  training of recurrent neural networks,'' \emph{IEEE Transactions on Neural
  Networks}, vol.~22, no.~4, pp. 588--600, 2011.

\bibitem{Haarnoja2016}
\BIBentryALTinterwordspacing
T.~Haarnoja, A.~Ajay, S.~Levine, and P.~Abbeel, ``{Backprop KF: Learning
  Discriminative Deterministic State Estimators},'' in \emph{Advances in Neural
  Information Processing Systems (NIPS)}, 2016. [Online]. Available:
  \url{http://arxiv.org/abs/1605.07148}
\BIBentrySTDinterwordspacing

\bibitem{Jonschkowski2018}
R.~Jonschkowski, D.~Rastogi, and O.~Brock, ``{Differentiable Particle Filters:
  End-to-End Learning with Algorithmic Priors},'' in \emph{RSS}, 2018.

\bibitem{karkus2018particle}
P.~Karkus, D.~Hsu, and W.~S. Lee, ``Particle filter networks with application
  to visual localization,'' \emph{arXiv preprint arXiv:1805.08975}, 2018.

\bibitem{Krishnan2017}
R.~G. Krishnan, U.~Shalit, and D.~Sontag, ``{Structured Inference Networks for
  Nonlinear State Space Models},'' in \emph{Association for the Advancement of
  Artificial Intelligence (AAAI)}, no. Dmm, 2017, pp. 1--21.

\bibitem{Fraccaro2016}
M.~Fraccaro, S.~K. S{\o}nderby, U.~Paquet, and O.~Winther, ``{Sequential Neural
  Models with Stochastic Layers},'' in \emph{Advances in Neural Information
  Processing Systems (NIPS)}, 2016.

\bibitem{Fraccaro2017}
M.~Fraccaro, S.~Kamronn, U.~Paquet, and O.~Winther, ``{A Disentangled
  Recognition and Nonlinear Dynamics Model for Unsupervised Learning},'' in
  \emph{Advances in Neural Information Processing Systems (NIPS)}, 2017.

\bibitem{Karl2016}
\BIBentryALTinterwordspacing
M.~Karl, M.~Soelch, J.~Bayer, and P.~van~der Smagt, ``{Deep Variational Bayes
  Filters: Unsupervised Learning of State Space Models from Raw Data},'' in
  \emph{International Conference on Learning Representations (ICLR)}, 2017.
  [Online]. Available: \url{http://arxiv.org/abs/1605.06432}
\BIBentrySTDinterwordspacing

\bibitem{Umlauft2017}
J.~Umlauft and S.~Hirche, ``{Learning Stable Stochastic Nonlinear Dynamical
  Systems},'' in \emph{International Conference on Machine Learning (ICML)},
  2017, pp. 3502----3510.

\bibitem{Davison2007}
A.~J. Davison, I.~D. Reid, N.~D. Molton, and O.~Stasse, ``{MonoSLAM: Real-Time
  Single Camera SLAM},'' \emph{IEEE Transactions on Pattern Analysis and
  Machine Intelligence}, vol.~29, no.~6, pp. 1052--1067, 2007.

\bibitem{Newcombe2011}
R.~A. Newcombe, S.~J. Lovegrove, and A.~J. Davison, ``{DTAM: Dense Tracking and
  Mapping in Real-Time},'' in \emph{IEEE International Conference on Computer
  Vision (ICCV)}, 2011, pp. 2320--2327.

\bibitem{Engel2014}
J.~Engel, T.~Sch{\"{o}}ps, and D.~Cremers, ``{LSD-SLAM: Large-Scale Direct
  Monocular SLAM},'' in \emph{European Conference on Computer Vision (ECCV)},
  2014.

\bibitem{Montiel2015}
R.~Mur-Artal, J.~Montiel, and J.~D. Tardos, ``{ORB-SLAM : A Versatile and
  Accurate Monocular SLAM System},'' \emph{IEEE Transactions on Robotics},
  vol.~31, no.~5, pp. 1147--1163, 2015.

\bibitem{Forster2015}
\BIBentryALTinterwordspacing
C.~Forster, L.~Carlone, F.~Dellaert, and D.~Scaramuzza, ``{IMU Preintegration
  on Manifold for Efficient Visual-Inertial Maximum-a-Posteriori Estimation},''
  in \emph{Robotics: Science and Systems}, 2015. [Online]. Available:
  \url{http://www.roboticsproceedings.org/rss11/p06.pdf}
\BIBentrySTDinterwordspacing

\bibitem{Li2013b}
M.~Li and A.~I. Mourikis, ``{High-Precision, Consistent EKF-Based
  Visual-Inertial Odometry},'' \emph{The International Journal of Robotics
  Research}, vol.~32, no.~6, pp. 690--711, 2013.

\bibitem{Bloesch2015}
M.~Bloesch, S.~Omari, M.~Hutter, and R.~Siegwart, ``{Robust Visual Inertial
  Odometry Using a Direct EKF-Based Approach},'' in \emph{IEEE International
  Conference on Intelligent Robots and Systems}, vol. 2015-Decem, 2015, pp.
  298--304.

\bibitem{Leutenegger2015}
S.~Leutenegger, S.~Lynen, M.~Bosse, R.~Siegwart, and P.~Furgale,
  ``{Keyframe-Based Visual–Inertial Odometry Using Nonlinear Optimization},''
  \emph{The International Journal of Robotics Research}, vol.~34, no.~3, pp.
  314--334, 2015.

\bibitem{Qin2018}
T.~Qin, P.~Li, and S.~Shen, ``{VINS-Mono: A Robust and Versatile Monocular
  Visual-Inertial State Estimator},'' \emph{IEEE Transactions on Robotics},
  vol.~34, no.~4, pp. 1004--1020, 2018.

\bibitem{Tang2019}
C.~Tang and P.~Tan, ``{BA-Net: Dense Bundle Adjustment Networks},'' in
  \emph{International Conference on Learning Representations (ICLR)}, 2019.

\bibitem{kashyap2020sparse}
H.~J. Kashyap, C.~C. Fowlkes, and J.~L. Krichmar, ``Sparse representations for
  object-and ego-motion estimations in dynamic scenes,'' \emph{IEEE
  Transactions on Neural Networks and Learning Systems}, 2020.

\bibitem{Mirowski2017}
\BIBentryALTinterwordspacing
P.~Mirowski, R.~Pascanu, F.~Viola, H.~Soyer, A.~J. Ballard, A.~Banino,
  M.~Denil, R.~Goroshin, L.~Sifre, K.~Kavukcuoglu, D.~Kumaran, and R.~Hadsell,
  ``{Learning to Navigate in Complex Environments},'' in \emph{International
  Conference on Learning Representations (ICLR)}, 2017. [Online]. Available:
  \url{http://arxiv.org/abs/1611.03673}
\BIBentrySTDinterwordspacing

\bibitem{Henriques2018}
J.~F. Henriques and A.~Vedaldi, ``{MapNet: An Allocentric Spatial Memory for
  Mapping Environments},'' in \emph{IEEE Conference on Computer Vision and
  Pattern Recognition (CVPR)}, 2018, pp. 8476--8484.

\bibitem{fragkiadaki2015recurrent}
K.~Fragkiadaki, S.~Levine, P.~Felsen, and J.~Malik, ``Recurrent network models
  for human dynamics,'' in \emph{Proceedings of the IEEE International
  Conference on Computer Vision}, 2015, pp. 4346--4354.

\bibitem{martinez2017human}
J.~Martinez, M.~J. Black, and J.~Romero, ``On human motion prediction using
  recurrent neural networks,'' in \emph{Proceedings of the IEEE Conference on
  Computer Vision and Pattern Recognition}, 2017, pp. 2891--2900.

\bibitem{tang2018long}
Y.~Tang, L.~Ma, W.~Liu, and W.-S. Zheng, ``Long-term human motion prediction by
  modeling motion context and enhancing motion dynamic,'' in \emph{Proceedings
  of the 27th International Joint Conference on Artificial Intelligence}, 2018,
  pp. 935--941.

\bibitem{shu2021spatiotemporal}
X.~Shu, L.~Zhang, G.-J. Qi, W.~Liu, and J.~Tang, ``Spatiotemporal co-attention
  recurrent neural networks for human-skeleton motion prediction,'' \emph{IEEE
  Transactions on Pattern Analysis and Machine Intelligence}, 2021.

\bibitem{Cimen2008}
T.~{\c{C}}imen, \emph{{State-Dependent Riccati Equation (SDRE) Control: A
  survey}}.\hskip 1em plus 0.5em minus 0.4em\relax IFAC, 2008, vol.~17, no. 1
  PART 1.

\bibitem{kingma2013auto}
D.~P. Kingma and M.~Welling, ``Auto-encoding variational bayes,''
  \emph{International Conference on Learning Representations (ICLR)}, 2014.

\bibitem{Rangapuram2018}
S.~S. Rangapuram, M.~Seeger, J.~Gasthaus, L.~Stella, Y.~Wang, and
  T.~Januschowski, ``{Deep State Space Models for Time Series Forecasting},''
  \emph{Advances in Neural Information Processing Systems}, no. NeurIPS, pp.
  7795--7804, 2018.

\bibitem{hochreiter1997long}
S.~Hochreiter and J.~Schmidhuber, ``Long short-term memory,'' \emph{Neural
  computation}, vol.~9, no.~8, pp. 1735--1780, 1997.

\bibitem{greff2016lstm}
K.~Greff, R.~K. Srivastava, J.~Koutn{\'\i}k, B.~R. Steunebrink, and
  J.~Schmidhuber, ``Lstm: A search space odyssey,'' \emph{IEEE transactions on
  neural networks and learning systems}, vol.~28, no.~10, pp. 2222--2232, 2016.

\bibitem{Schulman2015}
J.~Schulman, N.~Heess, T.~Weber, and P.~Abbeel, ``{Gradient Estimation Using
  Stochastic Computation Graphs},'' in \emph{Advances in Neural Information
  Processing Systems (NIPS)}, 2015, pp. 1--13.

\bibitem{Jankowiak2018}
\BIBentryALTinterwordspacing
M.~Jankowiak and F.~Obermeyer, ``{Pathwise Derivatives Beyond the
  Reparameterization Trick},'' in \emph{International Conference on Machine
  Learning (ICML)}, 2018. [Online]. Available:
  \url{http://arxiv.org/abs/1806.01851}
\BIBentrySTDinterwordspacing

\bibitem{Geiger2013}
A.~Geiger, P.~Lenz, C.~Stiller, and R.~Urtasun, ``{Vision Meets Robotics: The
  KITTI Dataset},'' \emph{The International Journal of Robotics Research},
  vol.~32, no.~11, pp. 1231--1237, 2013.

\bibitem{bian2019unsupervised}
J.~Bian, Z.~Li, N.~Wang, H.~Zhan, C.~Shen, M.-M. Cheng, and I.~Reid,
  ``Unsupervised scale-consistent depth and ego-motion learning from monocular
  video,'' in \emph{Advances in neural information processing systems}, 2019,
  pp. 35--45.

\bibitem{geiger2011stereoscan}
A.~Geiger, J.~Ziegler, and C.~Stiller, ``Stereoscan: Dense 3d reconstruction in
  real-time,'' in \emph{2011 IEEE intelligent vehicles symposium (IV)}.\hskip
  1em plus 0.5em minus 0.4em\relax Ieee, 2011, pp. 963--968.

\end{thebibliography}

\begin{IEEEbiography}
[{\includegraphics[width=1in,height=1.25in,clip,keepaspectratio]{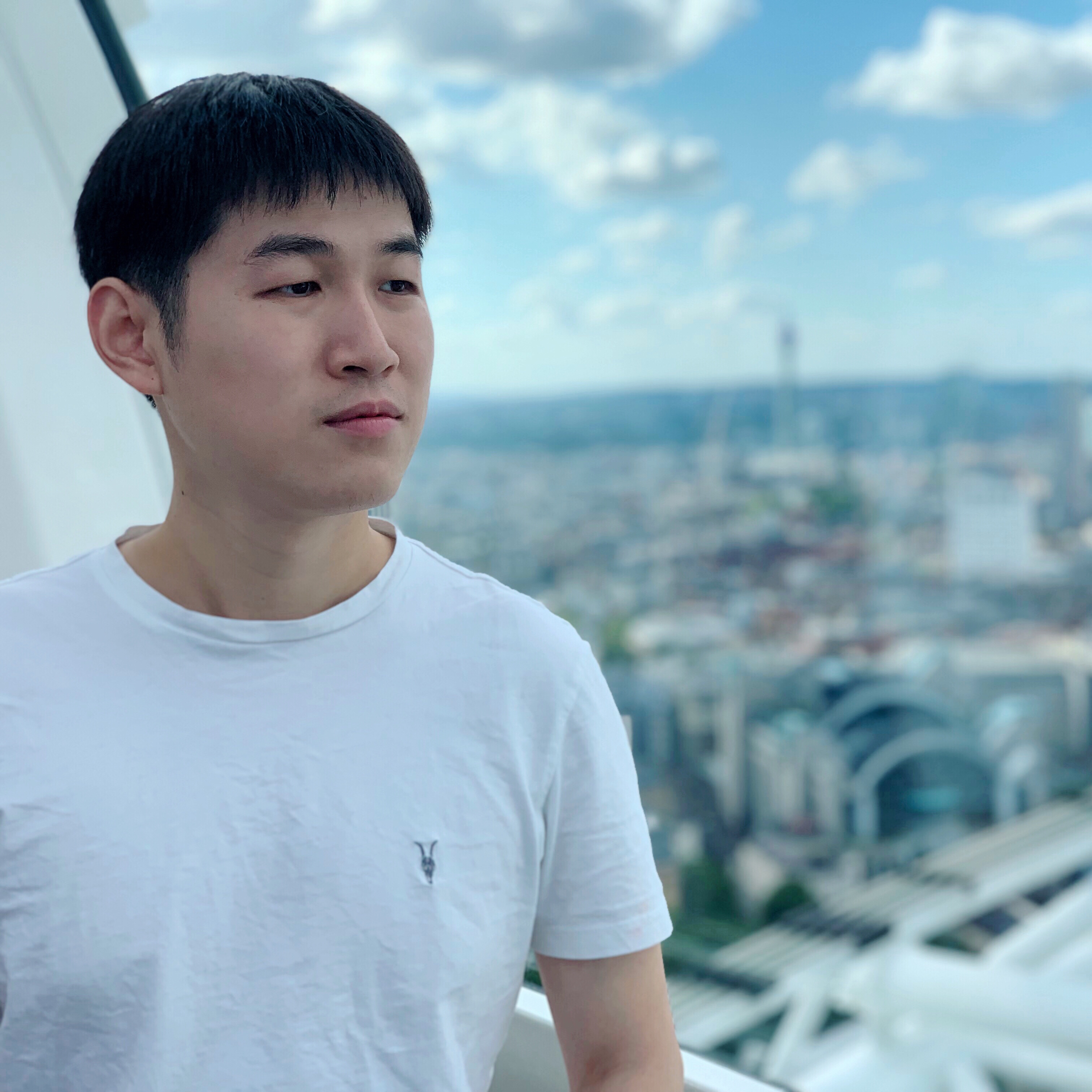}}]{Changhao Chen}
is currently a Lecturer at College of Intelligence Science, National University of Defense Technology (China).
Before that, he obtained his Ph.D. degree at University of Oxford (UK), M.Eng. degree at National University of Defense Technology (China), and B.Eng. degree at Tongji University (China). His research interest lies in robotics, machine learning and cyberphysical systems.
\end{IEEEbiography}

\vspace{-1.5cm}
\begin{IEEEbiography}
[{\includegraphics[width=1in,height=1.25in,clip,keepaspectratio]{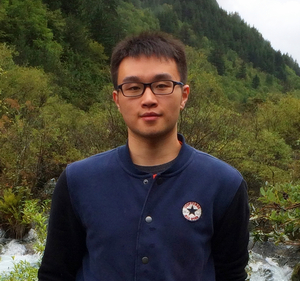}}]{Chris Xiaoxuan Lu}
is currently an Assistant Professor at School of Informatics, University of Edinburgh. Before that, he obtained his Ph.D degree at University of Oxford, and MEng degree at Nanyang Technology University,
Singapore. His research interest lies in Cyber Physical Systems, which use networked smart devices to sense and interact with the physical world.
\end{IEEEbiography}

\vspace{-1.5cm}
\begin{IEEEbiography}
[{\includegraphics[width=1in,height=1.25in,clip,keepaspectratio]{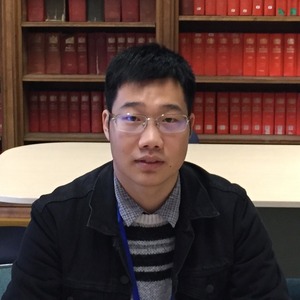}}]{Bing Wang}
 is currently PhD student at Department of Computer Science, University of Oxford. Before that, he obtained his BEng Degree at Shenzhen University, China. His research interest lies in camera localization, feature detection, description \& matching, and cross-domain representation learning.
\end{IEEEbiography}

\vspace{-1.5cm}
\begin{IEEEbiography}
[{\includegraphics[width=1in,height=1.25in,clip,keepaspectratio]{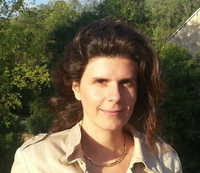}}]{Niki Trigoni}
is a Professor at the Department of Computer Science, University of Oxford. She is currently the director of the EPSRC Centre for Doctoral Training on Autonomous Intelligent
Machines and Systems, and leads the Cyber Physical Systems Group. Her research interests lie in intelligent and autonomous sensor systems with applications in positioning, healthcare, environmental monitoring and smart cities.
\end{IEEEbiography}

\vspace{-1.5cm}

\begin{IEEEbiography}
[{\includegraphics[width=1in,height=1.25in,clip,keepaspectratio]{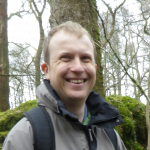}}]{Andrew Markham}
is an Associate Professor at the Department of Computer Science, University of Oxford. He obtained his BSc (2004)
and PhD (2008) degrees from the University of Cape Town, South Africa. He is the Director of the MSc in Software Engineering. He works on resource-constrained systems, positioning systems,
in particular magneto-inductive positioning and machine intelligence.
\end{IEEEbiography}

\end{document}